\documentclass{article} 
\usepackage{iclr2026_conference,times}
\usepackage{latexsym}

\usepackage{amsmath,amsfonts,bm}

\def\eqref#1{equation~\ref{#1}}

\def\1{\bm{1}}

\DeclareMathAlphabet{\mathsfit}{\encodingdefault}{\sfdefault}{m}{sl}
\SetMathAlphabet{\mathsfit}{bold}{\encodingdefault}{\sfdefault}{bx}{n}

\usepackage{hyperref}       
\usepackage{url}            
\usepackage{array}
\usepackage[dvipsnames]{xcolor}
\usepackage{xcolor}
\usepackage{tcolorbox}
\usepackage{booktabs} 
\usepackage{graphicx}
\usepackage{tabularx}
\usepackage{subcaption}
\usepackage[skip=2pt]{caption}
\usepackage{color, colortbl}
\usepackage{makecell}  
\usepackage{multirow}
\usepackage[normalem]{ulem}
\definecolor{blue1}{rgb}{0.215, 0.683, 0.844}
\definecolor{airforceblue}{rgb}{0.828,0.914,0.910}
\newcolumntype{P}[1]{>{\raggedright\arraybackslash}p{#1}}  
\hypersetup{
	colorlinks,
	citecolor=darkgray,
	linkcolor=blue1,
	urlcolor=blue}
\newcolumntype{L}[1]{>{\raggedright\arraybackslash}p{#1}}
\newcolumntype{R}[1]{>{\raggedleft\arraybackslash}p{#1}}

\usepackage[compact]{titlesec}
\titlespacing{\section}{0pt}{0.5ex}{0.5ex}
\titlespacing{\subsection}{0pt}{0ex}{0ex}
 \iclrfinalcopy
\author{\hspace{15ex} Siyuan Wang$^{*1,2}$\hspace{2ex} Gaokai Zhang$^{*1,3}$\hspace{2ex}   Li Lyna Zhang$^{1 \ddagger}$ \\\hspace{10ex}\bf   Ning Shang$^1$\hspace{4ex}  Fan Yang$^1$\hspace{4ex}   Dongyao Chen$^2$\hspace{4ex} Mao Yang$^1$ 
	\\\hspace{1ex}  \fontsize{10}{10} \selectfont{$^1$Microsoft Research Asia \hspace{5pt} $^2$Shanghai Jiao Tong University  	\hspace{5pt} $^3$Carnegie Mellon University}
}

\newcolumntype{g}{>{\columncolor{airforceblue}}c}
\newcommand{\sysname}{{LoongRL}}

\title{{\sysname}: Reinforcement Learning for Advanced Reasoning over Long Contexts}

\begin{document}

\newcommand{\lz}[1]{\textcolor{blue}{Lyna: #1}}
\maketitle
\def\thefootnote{$*$}\footnotetext{Equal contribution; work was done during the internship at Microsoft Research Asia}
\def\thefootnote{$\ddagger$}\footnotetext{Correspondence to lzhani@microsoft.com}

\begin{abstract}

Reasoning over long contexts is essential for large language models. While reinforcement learning (RL) enhances short-context reasoning by inducing "Aha" moments in chain-of-thought, the advanced thinking patterns required for long-context reasoning  remain largely unexplored, and high-difficulty RL data are scarce.
 In this paper, we introduce \textbf{\sysname}, a data-driven RL method for advanced long-context reasoning. Central to {\sysname} is \textit{KeyChain},  a synthesis approach that transforms short multi-hop QA into \textit{high-difficulty} long-context tasks by inserting UUID chains that hide the true question among large collections of distracting documents. Solving these tasks requires the model to trace the correct chain step-by-step, identify the true question,   retrieve relevant facts and reason over them to answer correctly. 
RL training on KeyChain data induces an emergent \textbf{plan–retrieve–reason–recheck} reasoning pattern that generalizes far beyond training length. Models trained at 16K effectively solve 128K tasks without prohibitive full-length RL rollout costs. On Qwen2.5-7B and 14B, {\sysname} substantially improves long-context multi-hop QA accuracy by +23.5\% and +21.1\% absolute gains. The  resulting {\sysname}-14B reaches a score of 74.2, rivaling much larger frontier models such as o3-mini (74.5) and DeepSeek-R1 (74.9). It  also improves long-context retrieval, passes all 128K needle-in-a-haystack stress tests, and preserves short-context reasoning capabilities.

\end{abstract}

\section{introduction}

Reasoning over long input contexts is a critical capability for large language models (LLMs), as many real-world tasks, from analyzing legal documents to debugging large codebases, require  integrating information across thousands of tokens. Recent advances, such as OpenAI o-series~\citep{o1} and DeepSeek-R1~\citep{deepseekr1}, show that reinforcement learning can improve reasoning by eliciting longer chain of thoughts (CoT) and emergent self-reflection. However, these methods mainly target short-context inputs and rely on model internal knowledge (e.g., math reasoning).
 In contrast, long-context reasoning requires both reasoning and the ability to retrieve and ground information from extensive external input contexts. 
 Although modern models support longer context windows~\citep{achiam2023gpt,longrope}, they excel mainly at retrieval, leaving reasoning over long documents a persistent challenge for real-world tasks~\citep{ling2025longreason}.

This work aims to bridge this gap by enabling long-context models to move beyond retrieval and acquire advanced reasoning capabilities.  Inspired by the short-context reasoning successes~\citep{deepseekr1,gandhi2025cognitive}, we hypothesize that the key lies in discovering and mastering thinking patterns specific to long-context reasoning. Since such patterns remain unclear, we adopt a reinforcement learning approach to investigate whether high-quality reasoning patterns can emerge.

However, we face significant challenges. First, effective RL training requires difficult long-context problems that cannot be  solved by retrieval alone. Questions must be sufficiently challenging to  trigger reasoning, require retrieving relevant information from long input contexts during reasoning, and have verifiable answers, as recent RL methods rely on outcome-only rewards to avoid reward hacking~\citep{deepseekr1,lambert2024tulu}. However, such data is extremely scarce, and answers often take multiple valid forms, making reliable verification difficult.  Second,  strong long-context performance typically requires training at near-target lengths~\citep{liu2024deepseek,li2025minimax}, but scaling RL rollouts from short inputs 
 (i.e., current $<$1K tokens) to 128K contexts incurs prohibitive compute and memory costs, making direct training infeasible at standard compute scales.
 Third, even if feasible,  training exclusively on long-context data risks degrading  short-context and general reasoning abilities~\citep{peng2023yarn,longrope2}, which remain critical  in practice. 

\begin{figure}[t]
	\centering
	\includegraphics[width=1\linewidth]{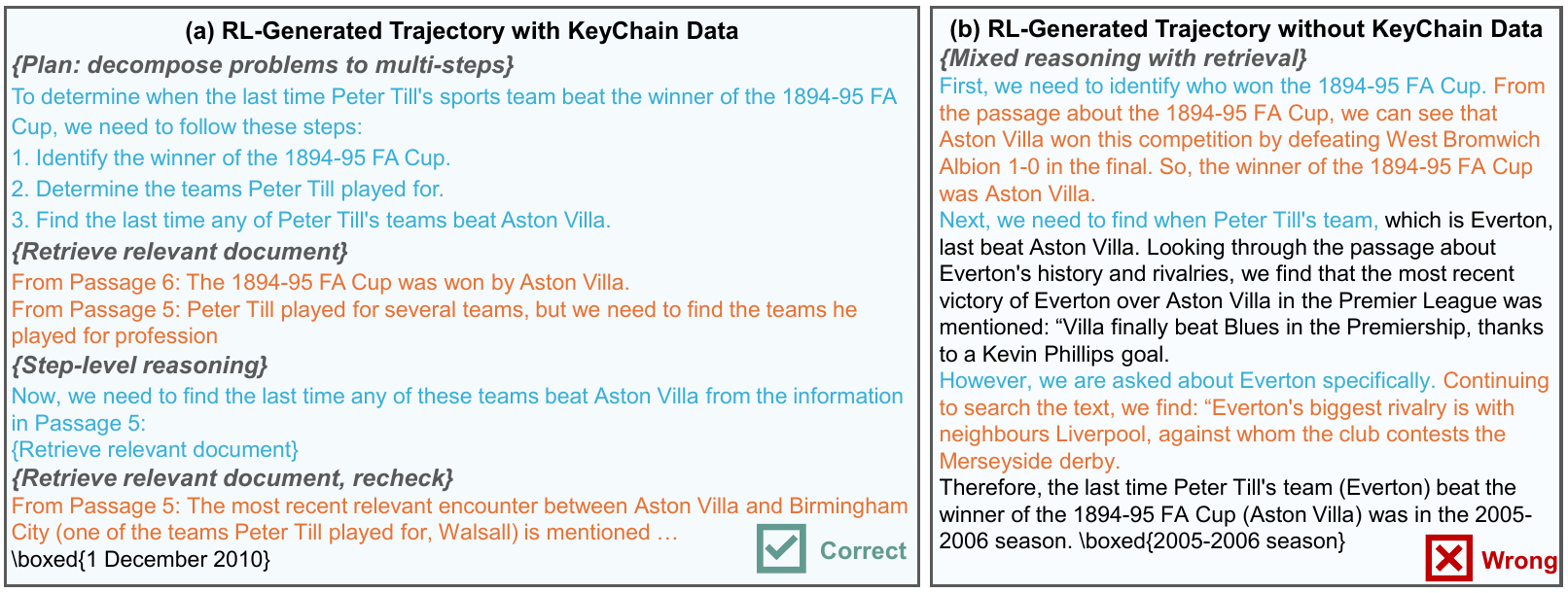}
	\caption{Model trajectories on long-context multi-hop QA with and without KeyChain RL data. \textbf{(a)} With KeyChain data, model exhibits an emergent plan–retrieve–reason–recheck thinking pattern, improving reasoning reliability and can generalize to longer contexts. \textbf{(b)} Without KeyChain data, reasoning and retrieval are entangled, the model often lacks an explicit planning step and does not deeply reason over  retrieved information, frequently leading to errors.   Reasoning steps are marked in blue and retrieval steps in orange. }
	\vspace{-4ex}
	\label{fig:reasoningpattern}
\end{figure}

To this end, we introduce \textbf{\sysname}, a data-driven reinforcement learning method that incentivizes models to  acquire effective thinking patterns for advanced long-context reasoning.  At its core is \textbf{KeyChain}, a synthesis approach that transforms short multi-hop QA datasets into \textit{high-difficulty} long-context problems by extending inputs with distracting documents and inserting UUID “chains” that hide the true question across multiple hops. Solving these problems requires the model to trace the correct chains step-by-step, identify the actual question, retrieve relevant facts from the long context, and reason over them to generate the answer. To enable reliable RL training, we design a rule-based answer verifier, \textit{two-way substring exact match}, which effectively evaluates free-form answers in general QA while mitigating reward hacking. Using KeyChain data, RL consistently elicits an emergent \textbf{plan–retrieve–reason–recheck} reasoning pattern, as shown in Fig.~\ref{fig:reasoningpattern}(a). Remarkably, this emergent patterns generalizes beyond the training length, enabling models trained at 16K to effectively handle 128K reasoning tasks without the prohibitive cost of full-length RL. Finally, we introduce a balanced data-mixing strategy to enhance long-context reasoning while preserving short-context general reasoning and long-context retrieval capabilities.

Extensive experiments across Qwen2.5-7B-Instruct and Qwen2.5-14B-Instruct and diverse  benchmarks demonstrate the superiority of {\sysname}. Remarkably, {\sysname} substantially boosts Qwen2.5-7B-Instruct and Qwen2.5-14B-Instruct by \textbf{+23.5\%} and \textbf{+21.1\%} absolute accuracy improvements on long-context multi-hop QA tasks. The resulting {\sysname} 14B achieves a score of 74.2, significantly surpassing all baselines and closely approaching much larger models such as o3 mini at 74.5 and DeepSeek-R1 at 74.9. Beyond the 16K training length, {\sysname} generalizes effectively up to 128K tokens, substantially improving long-context retrieval and passing all needle-in-a-haystack pressure tests.
At the same time, it preserves short-context and general reasoning capabilities, setting a new state of the art for models at this scale, and shows that {\sysname} can induce advanced reasoning patterns to substantially improve long-context reasoning.

\section{Related Works}

\noindent\textbf{Reasoning and Long-Context Reasoning}. Recent advances in LLM reasoning are largely driven by high-quality human-like chains of thought (CoT), typically obtained via teacher model distillation~\citep{qwen3} or self-generation through  reinforcement learning~\citep{deepseekr1}. Most existing studies focus on short-context reasoning tasks, such as mathematics~\citep{wang2025reinforcement,rstar2agent} and code~\citep{rstarcoder,ahmad2025opencodereasoning}, where emergent patterns like self-reflection and “aha” moments are crucial to the success~\citep{gandhi2025cognitive}. In contrast, exploration of advanced long-context reasoning patterns remains limited.

Existing efforts to improve long-context reasoning largely fall into two categories: prompting-based methods~\citep{yen2024helmet} and synthetic-data SFT~\citep{li-etal-2024-large,li-etal-2024-alr2,li-etal-2024-making}. Prompting is limited by the base model’s reasoning capacity, while synthetic-data SFT often introduces noise or bias, constraining advanced capabilities. QwenLong-L1~\citep{wan2025qwenlong} makes a notable step by extending R1-distill-Qwen-32B with RL on sequences up to 60K tokens, encouraging self-exploration of long reasoning trajectories. However, it leaves open key questions about how to design high-quality RL training data. We address this gap by introducing KeyChain RL data that fosters emergent reasoning patterns and generalizes from 16K to 128K contexts with significantly higher efficiency.

\noindent\textbf{Long-Context Synthetic Data}. Existing methods for long-context data synthesis primarily extend input contexts by padding questions with additional irrelevant documents. For example, \citet{li-etal-2024-making} augment MuSiQue~\citep{musique} with extra unrelated passages; \citet{li-etal-2024-alr2} use document-filling on HotpotQA~\citep{hotpotqa} and SQuAD~\citep{rajpurkar-etal-2016-squad}; and \citet{li-etal-2024-large} shuffle MuSiQue passages similarly. While these approaches increase context length, they are limited in generating high-quality, challenging training data.

\section{Methodology}

\begin{figure}[t]
	\centering
	\includegraphics[width=\textwidth]{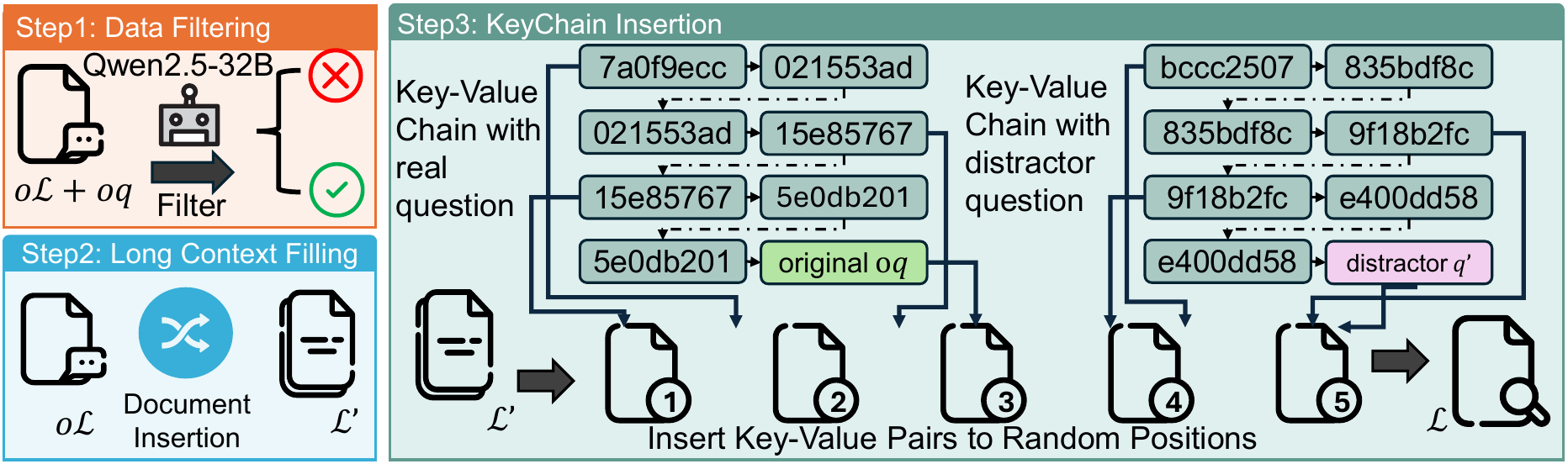}
	\caption{Overview of our KeyChain data construction.}
	\label{fig:data_paradigm}
\end{figure}

\subsection{KeyChain Data Construction}
\label{sec:keychain}

\noindent\textbf{Overview}. {\sysname} is a data-driven reinforcement learning approach designed to train models with advanced reasoning over long contexts. It relies on a high-quality RL training dataset  $\mathbb{D}=\{\mathcal{L}_i, q_i, a_i\}$ constructed under three principles: \textbf{(i)} each question $q_i$ and answer $a_i$ are  from real-world datasets to ensure reliability, as  synthetic data often suffer from hallucination~\citep{rstarcoder}; \textbf{(ii)} solving question $q_i$ requires reasoning over the full long input context $\mathcal{L}_i$, not merely leveraging model internal knowledge or direct retrieval. \textbf{(iii)} questions $q_i$ are sufficiently challenging to allow RL to incentivize advanced long-context reasoning capabilities. 

Fig.~\ref{fig:data_paradigm} illustrates the  KeyChain data construction. We begin with curated, high-quality short-context QA pairs $\{o\mathcal{L}_i,  {oq}_i, {oa}_i\}$  from real-world tasks. Each example is first 
expanded into a long input $\mathcal{L}'_i$ of $~$16K tokens by inserting distracting documents. KeyChain then transforms $\{\mathcal{L}'_i, {oq}_i, {oa}_i\}$ into $\{\mathcal{L}_i, q_i, a_i\}$ by randomly inserting multi-hop key-value chains that hide the original question ${oq}_i$ within ${\mathcal{L}_i}$, which significantly increases difficulty. 
 Given the new question $q_i$, model must first traces the chain to recover ${oq}_i$, and then perform long-context reasoning over $\mathcal{L}_i$ to generate the correct answer $a_i$, where $a_i={oa}_i$.  This construction ensures that RL training focuses on reasoning over long contexts rather than memorization or shallow retrieval.

\noindent\textbf{Seed Dataset Curation and Context Extension}.  We curate a high-quality seed dataset from three real-world multi-hop QA datasets: HotpotQA~\citep{hotpotqa}, MuSiQue~\citep{musique}, and 2WikiMultiHopQA~\citep{2wikimultihopqa}. Each question ${oq}_i$ is paired with its ground-truth answer ${oa}_i$ and requires reasoning across multiple documents within a short context $o\mathcal{L}_i$.
This initial collection contains 277K QA instances. To ensure effective RL training, we filter out tasks that are overly easy or excessively hard questions. Specifically, we answer each question eight times using Qwen2.5-32B-Instruct~\citep{qwen2.5}, and discard those  with a pass rate of 0 or 1. This yields 72K examples of moderate difficulty.

We then extend  each short context  $o\mathcal{L}_i$ into a long context $\mathcal{L}_i'$ by inserting additional real-world documents while keeping the original question ${oq}_i$ unchanged. The extra documents are sampled from the short-context documents of the 200K filtered-out QA tasks, excluding any overlap with $o\mathcal{L}_i$. Each extended context is approximately ($<$) 16,384 tokens, requiring the model to retrieve relevant information from a large set of distractors. This construction closely simulates real-world long-context reasoning, where relevant information is often buried within extensive irrelevant text.

\noindent\textbf{KeyChain Data Construction}. To make reinforcement learning effective for long-context reasoning, we build upon the above long-context multi-hop QA data to construct the KeyChain dataset.  Fig.~\ref{fig:data_paradigm} illustrates the  process. For each long-context QA task $\{\mathcal{L}'_i, {oq}_i, {oa}_i\}$, we insert linear key-value chains into the context, resulting in $\mathcal{L}_i$.  In each chain, a key maps to a value that contains the next key,   forming a step-by-step tracing path. We design two types of chains: \textit{(i)} one chain that ultimately resolves to the original question ${oq}_i$, and \textit{(ii)} multiple chains that instead resolve to distracting questions. Each key  is generated as a 32-character UUID, with character randomly sampled  from 0-9 and A-F. Distracting questions are randomly sampled from other QA instances in the dataset to ensure they are plausible but irrelevant. 

From  the first type of chain, we then construct  a new question $q_i$. This question requires the model to start from the initial key, trace the chain  within  $\mathcal{L}_i$, recover the original question ${oq}_i$, and finally perform long-context reasoning over $\mathcal{L}_i$ to produce the correct answer $a_i={oa}_i$. This design substantially increases task difficulty, as the model must first localize the hidden question and then reason over the extended context to answer correctly. An example of an augmented KeyChain long-context question is shown below:
\vspace{-1ex}
\begin{center}
	\label{self-correct}
	\fontsize{8}{8} \selectfont
\begin{tcolorbox}[%
	colback=white,
	colframe=gray!75!black,
	title=Example of KeyChain-augmented long-context question]
	\small\ttfamily
	Please read the following text. \\
	<Document 0> \\
	<original text>
	\textcolor{blue1}{\{"UUIDB‑n":
		"distracting question"\}} 	<original text>\\
	<Document 1> \\
	\textcolor{blue1}{\{"\textbf{UUIDA‑1}":
		"\textbf{UUIDA‑2}"\}} \\
	<Document 2> \\
	\textcolor{blue1}{\{"UUIDB‑1":
		"UUIDB‑2"\}} \\
	\ldots \\
	\textcolor{blue1}{\{"\textbf{UUIDA‑n}":
		"\textbf{correct question}"\}} \\
	\ldots \\
	In the context above, there is one correct question to answer. The correct question can only be found by following the correct consecutive chain of key:value pairs encoded with UUID strings (e.g., f81d4fae-7dec-11d0-a765-00a0c91e6bf6), starting from \texttt{\textcolor{blue1}{"starting UUIDA-1"}}.\\
	Find the correct question first, then answer it.
\end{tcolorbox}
\vspace{-1ex}
\end{center}
\noindent\textbf{Emergent Long-Context Reasoning Patterns}. We surprisingly find that RL training with KeyChain data enables models to develop emergent, human-like long-context reasoning patterns. As shown in Fig.~\ref{fig:reasoningpattern}(a), for each long-context QA task, the model first generates an explicit plan decomposing the problem into subproblems and substeps, retrieves relevant information for each step, and actively re-checks retrieved content when uncertain before proceeding. This structured \textbf{plan–retrieve–reason–recheck} loop leads to highly logical and reliable solutions.  Furthermore, we observe that this reasoning pattern also improves conventional long-context retrieval tasks, as illustrated in Appendix~\ref{sec:appretrieval} with an example trajectory on the RULER vt benchmark, where the model performs step-by-step, human-readable retrieval, progressively locating the correct answer rather than directly jumping to it as in traditional retrieval approaches.

More importantly, the plan–retrieve–reason–recheck behavior learned on short contexts (16K tokens) \textbf{generalizes to much longer contexts}, up to 128K tokens (see Experiments). This allows training on 16K sequences while maintaining strong longer context performance, highlighting the robustness and scalability of the KeyChain RL approach.

\subsection{Long-Context Reinforcement Learning}

This section introduces our long-context reinforcement learning methodology using KeyChain data, covering reward design, data mixing and multi-stage training recipe. 

\subsubsection{Group Relative Policy Optimization for Long-Context Reasoning}
\label{sec:grpo}
\noindent\textbf{Group Relative Policy Optimization (GRPO)}. For training, we adopt the GRPO algorithm.  Specifically, 
for each question $q$, its long context $\mathcal{L}$, and its ground-truth answer $a$ from a dataset ${D}$, GRPO samples a group of rollout trajectories $\{o_1,o_2,\cdots,o_{G}\}$ from the old policy $\pi_{\theta_{old}}$ and then optimizes the policy  $\pi_{\theta}$ by maximizing the following objective:
\begin{align}
	J_{\text{GRPO}}(\theta) 
	&= \mathbb{E}_{(\mathcal{L}, q, a) \sim \mathcal{D},\, \{o_i\}_{i=1}^{G} \sim \pi_{\theta_{\text{old}}}(\cdot|q)} \notag \\
&	\Bigg[ \frac{1}{G} \sum_{i=1}^{G} \frac{1}{|o_i|} \sum_{t=1}^{|o_i|} 
	\Big( \min \big[ \rho_{i,t}(\theta) A_{i,t}, 
	\mathrm{clip}(\rho_{i,t}(\theta), 1-\varepsilon, 1+\varepsilon) A_{i,t} \big]
	\quad - \beta D_{\text{KL}}(\pi_\theta \| \pi_{\text{ref}}) \Big) \Bigg]
\end{align}
where $	\rho_{i,t}(\theta) = \frac{\pi_\theta(o_{i,t} | q, o_{i,<t})}{\pi_{\theta_{\text{old}}}(o_{i,t} | q, o_{i,<t})}$. 
Hyper-parameters $\varepsilon$ and $\beta$ control the clipping range of importance sampling ratio and the weight of KL penalty term, respectively. The estimated advantage 
$A_{i,t}$ is computed from a group of rewards $\{r_1,r_2,...r_G\}$ for each rollout trajectory:
\begin{align}
A_{i,t}=\frac{r_i-\text{mean}(\{r_1,r_2,\cdots,r_G\})}{\text{std}(\{r_1,r_2,\cdots,r_G\})}
\end{align}
Here, $r_i$ is the reward for trajectory $o_i$, evaluated using a rule-based verifier to mitigate reward hacking
~\citep{deepseekr1,kimi1.5}.  

To stabilize RL training, we follow best practices. A small KL penalty $\beta=0.001$ prevents excessive policy deviation. Following prior works~\citep{rstar2agent}, we  remove the entropy loss term, which while commonly used to encourage exploration, can cause uncontrolled entropy growth and destabilize training, so it is omitted in our experiments.

\noindent\textbf{Rule-based Reward Design}. In our long-context RL, most questions are general QA rather than math or code problems with clear answers. The answers can take many valid forms, making it difficult to determine whether a rollout trajectory truly reaches the correct solution. 
 Prior works such as QwenLong-L1~\citep{wan2025qwenlong} address this by using LLM-as-a-judge, but this introduces additional complexity. In addition to the already expensive long-context RL training, it requires serving another model for answer judgment, while still leaving room for reward hacking. 

We instead adopt a rule-based reward, following the success of verifiable rewards in mathematical and code RL~\citep{rstar2agent,wang2025beyond,deepseekr1}. Our approach is simple yet effective for  long-context reasoning. First, we explicitly require the model to output its final answer within \textbackslash\texttt{boxed}\{\} in the training prompt (in Appendix.~\ref{appendix:training_prompt}), ensuring unambiguous answer extraction. Second, we apply a \textit{two-way substring exact match} on the boxed answer. Each rollout trajectory $o_i$ receives a binary accuracy reward $r_i\in \{0,1\}$ depending on whether the extracted final answer $y_\text{ans}$ contains the ground truth answer $a$ as a substring, or the ground truth answer $a$ contains $y_\text{ans}$ as a substring. Formally, the reward is computed as:
\begin{align}
	\label{eq:reward}
	\quad r_{i}=\begin{cases} 1 &\text{if $\{a \subseteq y_{\text{ans}} \;\lor\; y_{\text{ans}} \subseteq a\}$}, \\ 0 & \text{otherwise}. \end{cases}
\end{align}
Compared to strict exact match, this design tolerates valid answer variations and avoids the rigidity that may otherwise exclude correct outputs. Experiments in Table~\ref{tbl:verifier} demonstrate its effectiveness.

\subsubsection{Training Recipe}

We conduct {\sysname} training on Qwen2.5-7B-Instruct and Qwen2.5-14B-Instruct, both with a 128K context window. The goals are (i) enhancing long-context reasoning through reinforcement learning, and (ii) preserving core abilities such as general short-context reasoning. To achieve this, we construct a mixed dataset (Table~\ref{tab:datasets}) and adopt a multi-stage RL training strategy.

\begin{table}[t]
	\centering
	\caption{Data recipe for long-context RL training.}
	\label{tab:datasets}
	\small
	\resizebox{0.98\textwidth}{!}{
	\begin{tabular}{@{}llrrr@{}}
		\toprule
		\textbf{Dataset} & \textbf{Description} & \textbf{\# Size} & \textbf{Length range} &Difficulty\\
		\midrule
		HotpotQA-KeyChain & KeyChain-augmented HotpotQA & 2,500 & 16,272--20,670 & Hard\\
		MuSiQue-KeyChain & KeyChain-augmented MuSiQue & 2,500 & 16,495--20,623 &Hard\\
		2WikiMultiHopQA-KeyChain & KeyChain-augmented 2WikiMultiHopQA & 2,500 & 14,911--20,576 &Hard\\
		HotpotQA & Standard multi-hop QA & 2,500 & 12,058--16,279 &Medium\\
			MuSiQue & Standard multi-hop QA & 2,500 & 12,562--16,283 &Medium\\
				2WikiMultiHopQA & Standard multi-hop QA & 2,500 & 10,727--16,274 &Medium\\
				\midrule
		Book RULER (Multi-key) & Long-context retrieval (20 keys, 1 value) & 512 & 12,038--17,387 &Easy\\
		Book RULER (Multi-value) & Long-context retrieval (1 key, 20 values) & 512 & 11,648--17,840 &Hard\\
		\midrule
		Math Choice & Multiple-choice math problems & 2,500 & 40--425 & Easy \\
			DAPO Math & Mathematical reasoning & 2,500 & 65--1,014 & Hard\\
		\bottomrule
	\end{tabular}}
\vspace{-2ex}
\end{table}
\noindent\textbf{Training Length and Data Mix}. As discussed in Sec.~\ref{sec:keychain}, KeyChain data effectively induces long-context reasoning patterns, enabling the model to generalize to longer contexts. To avoid the high cost of full 128K RL rollouts, we train using a 16K context length.

Table~\ref{tab:datasets} summarizes our training data sources, including their input context lengths and task difficulty. Our dataset consists of four types. \textit{(i) High-difficulty KeyChain data} is synthesized as described in Section~\ref{sec:keychain}, with 2,500 examples each from HotpotQA~\citep{hotpotqa}, MuSiQue~\citep{musique}, and 2WikiMultiHopQA~\citep{2wikimultihopqa}, totaling 7,500 examples. This set provides challenging examples that explicitly induce long-context reasoning. \textit{(ii) Medium-level multi-hop QA data} consists of 2,500 examples from each of the same three datasets. These moderately difficult examples are especially important for smaller models (e.g., Qwen2.5-7B-Instruct), enabling effective RL when the model initially struggles with harder KeyChain tasks. \textit{(iii) Long-context needle retrieval data} contains 1,024 synthetic examples designed to maintain the model’s ability to retrieve relevant information from long contexts. Each example uses a 16K-token book from PG19 as the base, into which multiple key–value “needles” are randomly inserted following RULER\citep{ruler}, requiring the model to locate relevant values amid extensive distractors. \textit{(iv) Math data} contains 5,000 short-context problems ($<$1K tokens) to preserve general short-context reasoning capabilities, including 2,500 hard problems from the DAPO training set~\citep{dapo_math} and 2,500 easy multiple-choice questions from MATH~\citep{hendrycks2021math}.

\noindent\textbf{Multi-Stage Training}. Our reinforcement learning follows a three-stage curriculum.
\textit{(i)} \textit{Warm-up.} We first train for one epoch on the dataset excluding KeyChain data. Since KeyChain problems are initially too difficult for small models, this stage allows the model to  improve retrieval and general reasoning ability on  easier data, ensuring stable optimization.
\textit{(ii)} \textit{Stage I (KeyChain augmentation).} KeyChain data is then introduced to increase task difficulty, encouraging the model to plan effectively, retrieve precise information from distractor-heavy long contexts, and integrate evidence into coherent reasoning chains.
\textit{(iii)} \textit{Stage II (difficulty-focused training).} After Stage I, we generate eight rollouts per example using the best checkpoint. Examples solved correctly in all rollouts are discarded, leaving a challenging subset ($~$30–40\% of the data). RL continues on this subset, focusing updates on difficult cases to improve efficiency while avoiding overtraining on mastered problems.

\section{Experiments}

\subsection{Setup}
\label{sec:setup}
\noindent\textbf{Training Setup.}  We run experiments on two long-context instruction-tuned models, \textit{Qwen2.5-7B-Instruct} and \textit{Qwen2.5-14B-Instruct}. Training uses GRPO using a group size $G=8$ and a learning rate of 1e-6. Batch sizes are set to 512 for 7B model and 256 for 14B model.  Rollouts are sampled with temperature $0.6$ and top-$p=0.95$, with a maximum output length of $4{,}096$ tokens and long-context inputs of $\sim16$K. We adopt a learning rate of $1\times10^{-6}$ with cosine decay and gradient clipping at $1.0$. For Qwen2.5-7B-Instruct, we apply the full three-stage RL training:  42 steps in warm-up, 168 in Stage I and 118 in Stage II.   For the larger Qwen2.5-14B-Instruct, we skip  warm-up stage since the model already possesses strong base abilities and can immediately handle KeyChain data. We train for $168$ steps in Stage~I and $150$ steps in Stage~II. 
For the 7B model, we train on 16×A100 GPUs, while the 14B model is trained on 8×MI300X GPUs.

\noindent\textbf{Evaluation Benchmarks}. We evaluate {\sysname}  models across three dimensions. \textbf{(i) Long-context reasoning}: we follow QwenLong-L1~\citep{wan2025qwenlong} and evaluate on multi-hop QA tasks in LongBench v1~\citep{longbench1},  including HotpotQA~\citep{hotpotqa}, 2WikiMultiHopQA~\citep{2wikimultihopqa}, MuSiQue~\citep{musique}), NarrativeQA~\citep{narrativeqa} and QASPER~\citep{qasper}, with input lengths  from 4K to 64K tokens.  We also evaluate on LongBench v2~\citep{longbench2}, a representative long-context reasoning benchmark supporting up to 128K tokens.  Due to space limits, detailed v2 results are in  Appendix Table~\ref{tab:lb2}. \textbf{(ii) General short-context reasoning}:  we use standard benchmarks including MMLU~\citep{hendrycks2020mmlu}, MATH-500~\citep{math-500}, and the instruction-following benchmark IFEval~\citep{zhou2023instruction}. \textbf{(iii) Long-context retrieval}: to measure the impact of long-context RL on retrieval abilities, we evaluate on  Needle in a Haystack~\citep{needlehaystack} and RULER~\citep{ruler}.

For inference, reasoning models and our models use temperature 0.6, with up to 128K input tokens and 10K output tokens. We sample eight solutions per problem and report average pass@1 accuracy. Non-reasoning models (e.g., Qwen2.5-7B-Instruct) use temperature 0.

\noindent\textbf{Baselines}. We compare against three  baselines: \textbf{(i)} leading frontier models, including o3-mini, GPT-4o, DeepSeek-R1 and QWQ-32B; \textbf{(ii)}
state-of-the-art  models enhancing short-context reasoning on long-context foundations, mainly R1-distilled variants; and \textbf{(iii)} long-context reasoning models like the recent QwenLong-R1-32B, based on R1-distill-Qwen-32B and trained with 60K input context.

\subsection{Main Results}

\begin{table}[t]
	\small
	\centering
	\caption{Results of {\sysname} and frontier LLMs on  long-context reasoning and general short tasks. {\sysname} delivers frontier-level long-context reasoning at much smaller scales (7B/14B), rivaling o3-mini and DeepSeek-R1, while preserving general short-context abilities across all scales. }
	\label{tab:main_results1}
	\setlength{\tabcolsep}{5pt}
	\resizebox{0.98\textwidth}{!}{	\begin{tabular}{@{\hskip0pt}l@{\hskip4pt}g@{\hskip2pt}c@{\hskip2pt}c@{\hskip4pt}c@{\hskip4pt}c@{\hskip4pt}c@{\hskip4pt}c@{\hskip4pt}g@{\hskip4pt}c@{\hskip4pt}c@{\hskip4pt}c@{\hskip0pt}}
			\toprule
			\multirowcell{2}{Models} & \multicolumn{6}{c}{Long-Context Reasoning}&& \multicolumn{4}{c}{General \& Short Reasoning}\\
			\cmidrule{2-7}  	\cmidrule{9-12}
			& {Avg.} & {HotpotQA} & {2WikiMultiHopQA} & { MuSiQue} & {NarrativeQA} & {QASPER} &&Avg.& {MMLU} & MATH&IFEval \\
			\midrule
			o3\textendash mini  (medium) & 74.5  & 83.0 & 89.0 & 64.0 & 60.7 & 60.5 &&\underline{\bf92.1} &86.9&\underline{\bf98.0}&\underline{\bf91.5} \\
			DeepSeek- R1            &\underline{\textbf{74.9}}    & 82.7 & 91.3 & \underline{\textbf{72.2}} &\underline{\textbf{66.9}} & 61.4& &90.5 &\underline{\bf90.8}&97.3&83.3 \\
			GPT\textendash 4o    & 64.7  & 82.5 & 78.0 & 54.0 & 60.5 & 48.5 & &82.5& 88.7&74.6&84.3 \\
			QwQ-32B                & 69.6     & 78.5 & 87.4 & 62.7 & 61.1 & 58.5 & &85.9 &75.7&\underline{\bf98.0}&83.9 \\
			R1-Distill-LLaMa-70B & 65.4& 76.1 & 85.0 & 61.9 & 53.4 & 50.5 & &85.4 &82.4&94.5&79.3 \\
			\midrule
			Qwen2.5-7B-Instruct  & 48.9 & 69.5 & 50.5 & 34.0 & 44.5 & 46.0 & &73.5 &73.4&76.0&\bf71.2 \\
			R1-Distill-Qwen-7B  & 31.2  & 40.2 & 53.3 & 11.1 & 8.9  & 42.5 &&69.9 &62.3&\bf92.8&54.7 \\
			\bf{\sysname}-7B& \bf 72.4&\underline{ \bf 83.1} & \bf91.1 & \bf65.6 & \bf58.4 & \bf63.6&  &\bf75.0 &\bf76.2&78.0&70.9 \\	
			\midrule 
			Qwen2.5-14B-Instruct & 53.1 & 74.0 & 60.5 & 36.5 & 48.5 & 46.0 & &81.3 &79.4&83.4&\bf 81.0 \\
			R1-Distill-Qwen-14B & 64.9  & 77.5 & 87.0 & 58.0 & 51.0 & 51.0 &&81.0 &76.6&93.9&72.6\\
			R1-Distill-Qwen-32B & 65.5 & 76.3 & 87.6 & 59.8 & 52.7 & 50.9  &&82.4&\bf 80.5&94.3& 72.5 \\
			QwenLong-L1-32B     & 70.1  & 80.7 & 89.1 & 65.2 & 58.6 & 56.7&  &\bf 84.1 &78.5&\bf95.2&78.6 \\
			\bf{\sysname}-14B& \bf74.2 & \bf82.2 &\underline{ \textbf{93.3}} & \bf67.5 & \bf63.4 & \underline{\textbf{64.5}}  &&80.7 &\bf 80.5&83.2&78.4 \\			
			\bottomrule	\end{tabular}}
			\vspace{-2ex}
\end{table}

\noindent\textbf{Competitive long-context reasoning at smaller scale}.  Table~\ref{tab:main_results1} summarizes the long-context reasoning performance of {\sysname} against state-of-the-art models. We highlight two key observations: \textbf{(i)} {\sysname} delivers frontier-level long-context reasoning  at significantly smaller scales. Remarkably, {\sysname}-7B achieves an average of 72.4 on LongBench v1, surpassing all R1-distilled models and QwenLong-L1-32B. At 14B, {\sysname} reaches 74.2, even rivaling the much larger, heavily trained  o3-mini (74.5) and DeepSeek-R1 (74.9). \textbf{(ii)} Our KeyChain-driven RL  proves  far more effective than existing methods. It improves Qwen2.5-7B-Instruct and Qwen2.5-14B-Instruct by \textbf{+23.5\%} and \textbf{+21.1\%}, respectively. In contrast, R1-distilled Qwen models, trained on long-CoT reasoning data, yield a modest +11.8\% gain at 14B and even degrade 7B performance by -17.7\%. Similarly, QwenLong-L1-32B, trained via conventional long-context RL on R1-distill-Qwen-32B, improves by just +4.6\% on average. Notably,  {\sysname}-7B even outperforms QwenLong-L1-32B by +2.3\%, demonstrating that much smaller models can surpass larger baselines with our approach.

\begin{table}[t]
	\small
	\centering
	\caption{While being trained only on 16K, {\sysname} generalizes impressively to context up to 128K.}
	\label{tab:lengthgeneralize}
	\begin{tabular}{ccccccccc}
		\toprule 
		\multirowcell{2}{Models} &\multicolumn{3}{c}{NarrativeQA} && \multicolumn{4}{c}{RULER}\\
		\cmidrule{2-4}\cmidrule{6-9}
		& 0-16K&16K-32K&32K-64K&& 16K&32K&64K&128K\\
		\midrule 
		Qwen2.5-7B-Instruct & 55.7& 35.2&  42.4& &92.3&89.5&81.8&69.4\\
		R1-Distill-Qwen-7B& 55.7& 35.2&42.4 & &18.9& 4.4&1.4&0.9\\
			\bf {\sysname}-7B & \bf69.8& \bf 47.4& \bf 57.2& &\bf 93.4& \bf 91.4& \bf 86.2& \bf 76.8\\
		\midrule 
		Qwen2.5-14B-Instruct& 55.7&40.7& 48.3& &93.4&92.5&82.3&73.6\\
		R1-Distill-Qwen-14B &63.0 &35.9&54.6&&85.7&82.0&60.2&28.2\\
		R1-Distill-Qwen-32B& 57.4& 44.4 & 58.9& & 90.3&88.9& 71.5& 40.9\\
		QwenLong-L1-32B & 65.9& 48.1 &60.0 & & 87.6& 86.8& 80.6& 70.2\\
\bf {\sysname}-14B & \bf 69.5 &\bf  55.2 & \bf 64.3& &\bf  95.4& \bf 95.1& \bf 87.1& \bf 79.9\\
		\bottomrule
	\end{tabular}
	\vspace{-2ex}
\end{table}

\noindent\textbf{Training at short, generalize better to long}. The strong results in Table~\ref{tab:main_results1} are  largely driven by KeyChain data, which enables our models to acquire a plan-retrieve-reason-recheck thinking pattern. Although trained  on 16K input contexts, this patterns generalizes effectively to much longer contexts. 
 As shown in Table~\ref{tab:lengthgeneralize},  both  {\sysname}-7B and 14B achieve substantial gains on longer-context reasoning and retrieval benchmarks. 
On NarrativeQA (32K-64K), they achieve impressive absolute gains of +14.8\% and +16.0\%, respectively, far exceeding R1-distilled models and QwenLong-L1-32B, which are trained with much longer contexts. 
On the RULER benchmark (up to 128K), while other baselines degrade sharply with increasing context length,  our models maintain consistently strong performance, showing that the learned reasoning pattern transfers robustly to longer contexts.

\noindent\textbf{Near-lossless general short reasoning}. Table~\ref{tab:main_results1} also reports {\sysname}’s performance on short-context reasoning and general tasks, showing that it effectively preserves the base models' capabilities. On MMLU, {\sysname}  even yields gains of \textbf{+2.8\%} (7B) and \textbf{+1.1\%} (14B). In contrast, both R1-distilled models and QwenLong-L1-32B suffer performance drops. On instruction following (IFEval), R1-distilled models degrade sharply (-16.5\% at 7B, -8.4\% at 14B), while {\sysname} shows only minimal declines (-0.3\% and -2.6\%). For math reasoning, although R1-distilled models benefit from heavy long-CoT data distillation, our approach stably preserves the base models’ math ability.

\begin{figure}[t]
	\centering
	\begin{subfigure}[t]{0.49\linewidth}
		\centering
		\includegraphics[width=\linewidth]{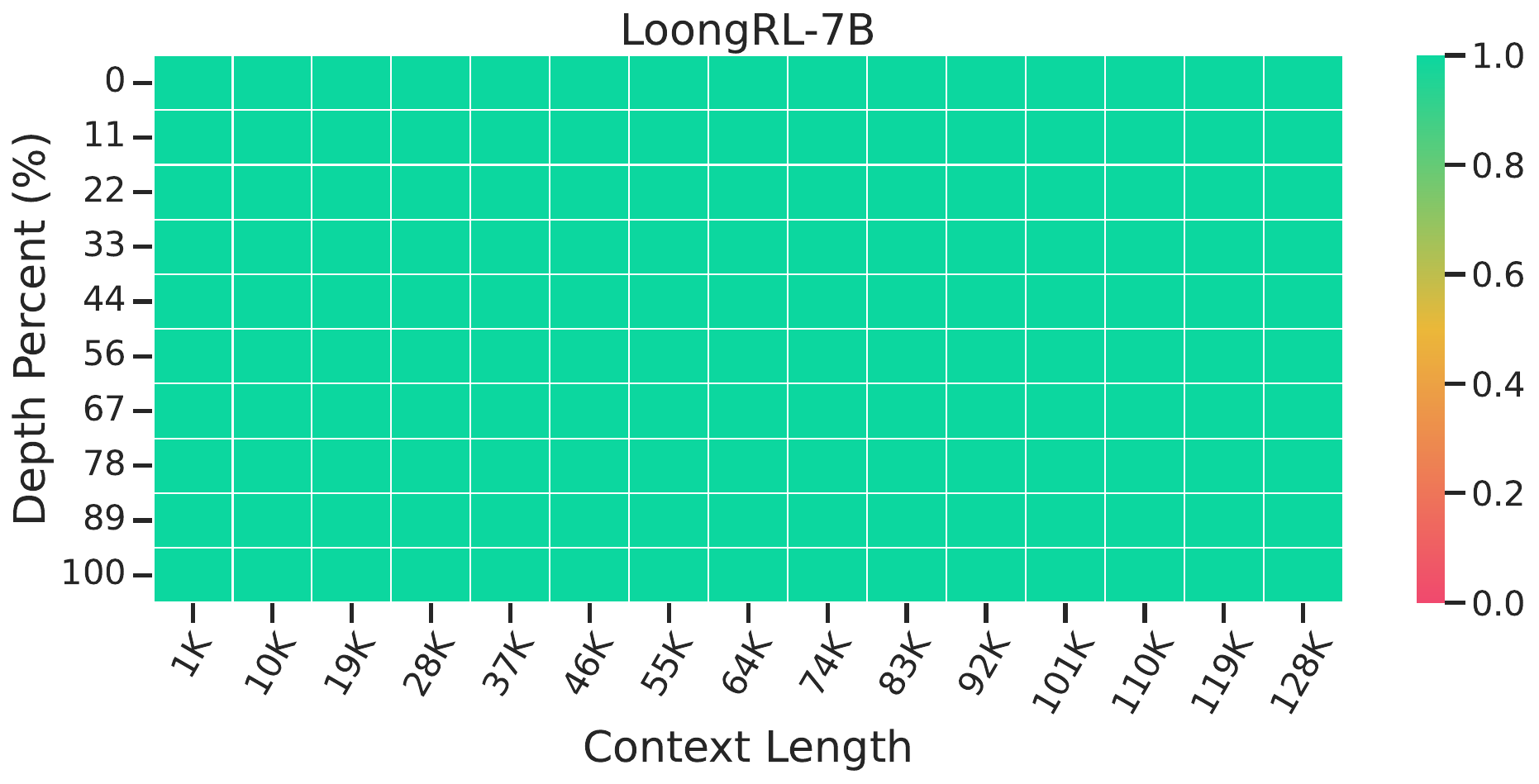}
	\end{subfigure}\hspace{0.01\linewidth}
	\begin{subfigure}[t]{0.49\linewidth}
		\centering
		\includegraphics[width=\linewidth]{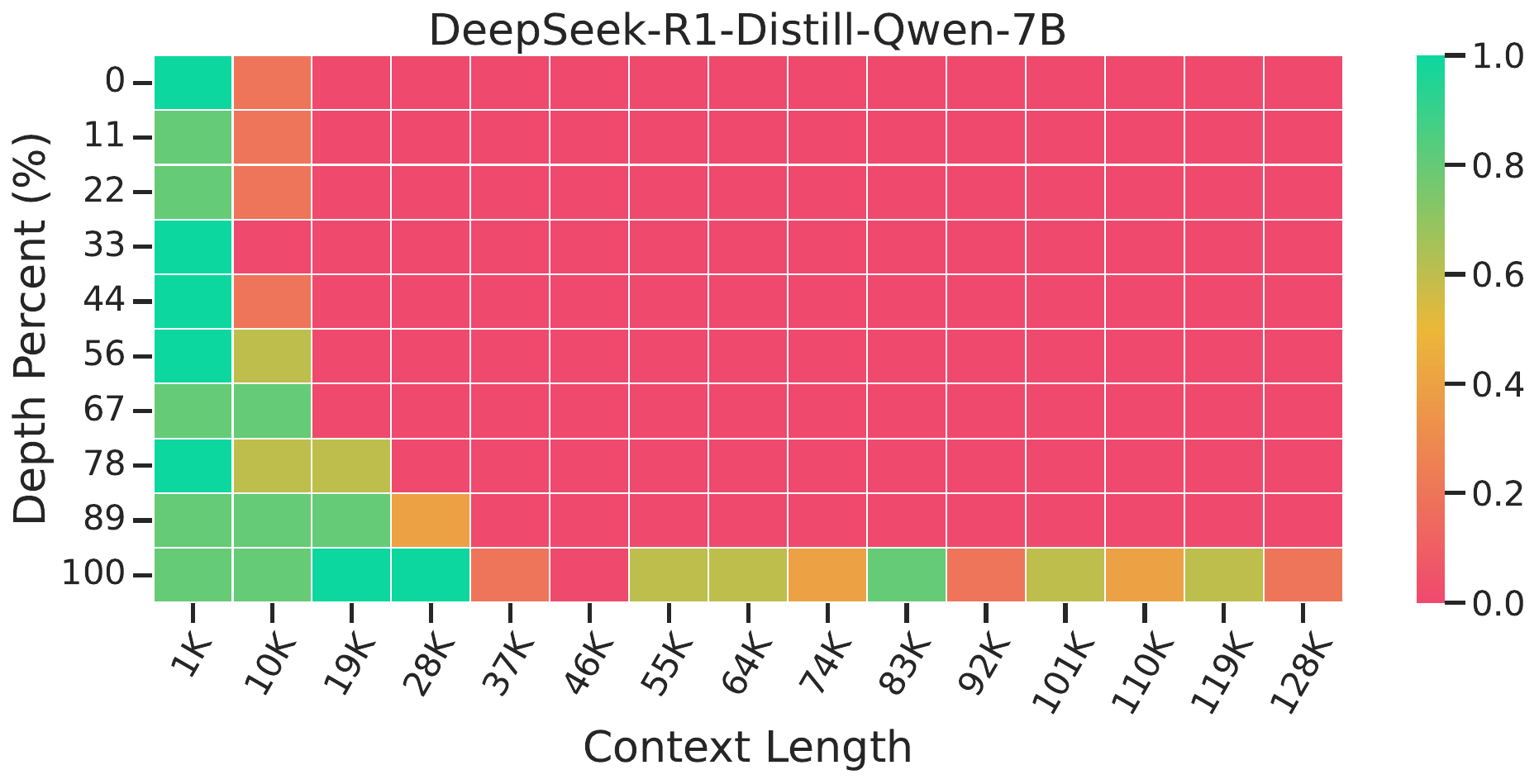}
	\end{subfigure}
	
	\begin{subfigure}[t]{0.49\linewidth}
		\centering
		\includegraphics[width=\linewidth]{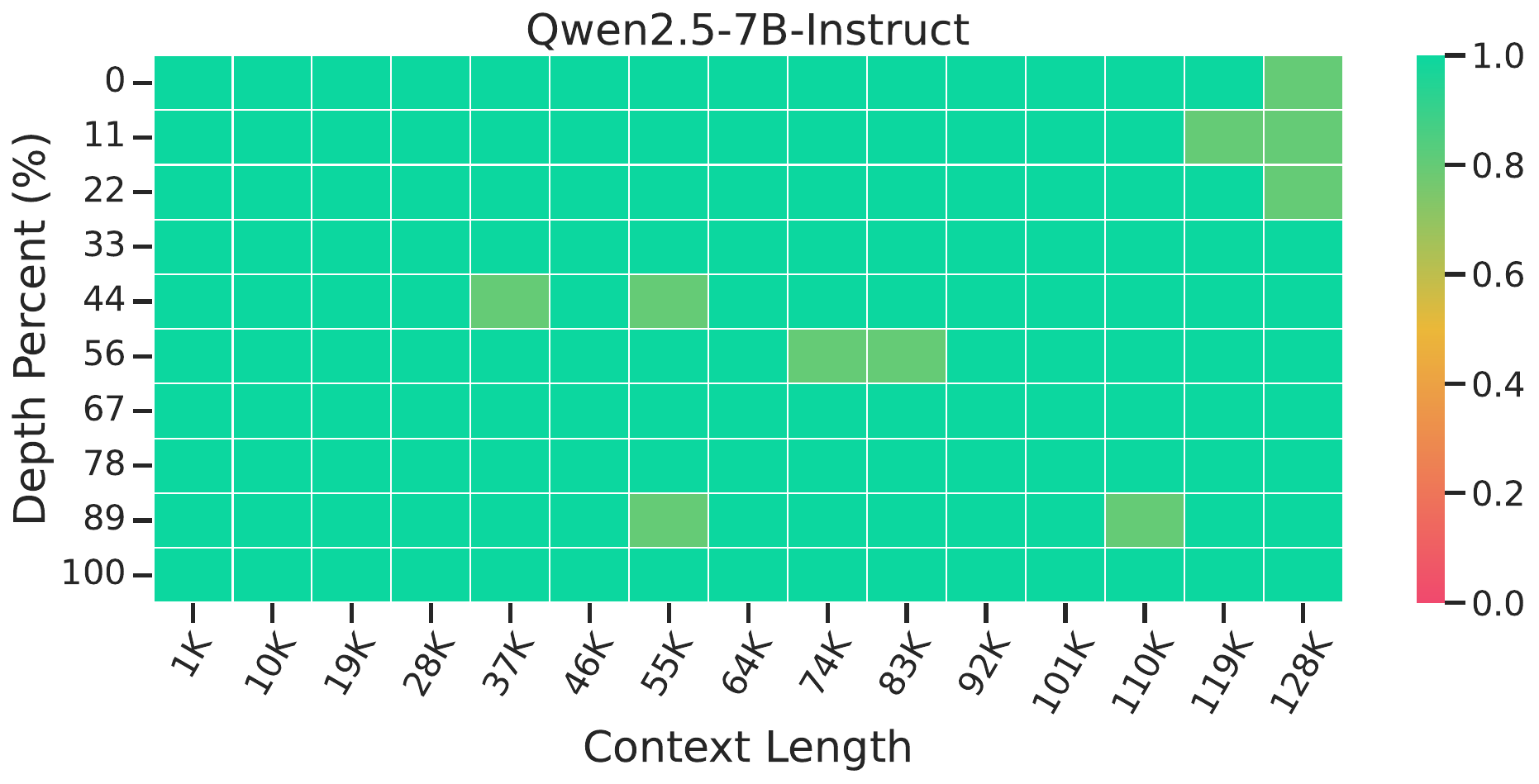}
	\end{subfigure}\hspace{0.01\linewidth}
	\begin{subfigure}[t]{0.49\linewidth}
		\centering
		\includegraphics[width=\linewidth]{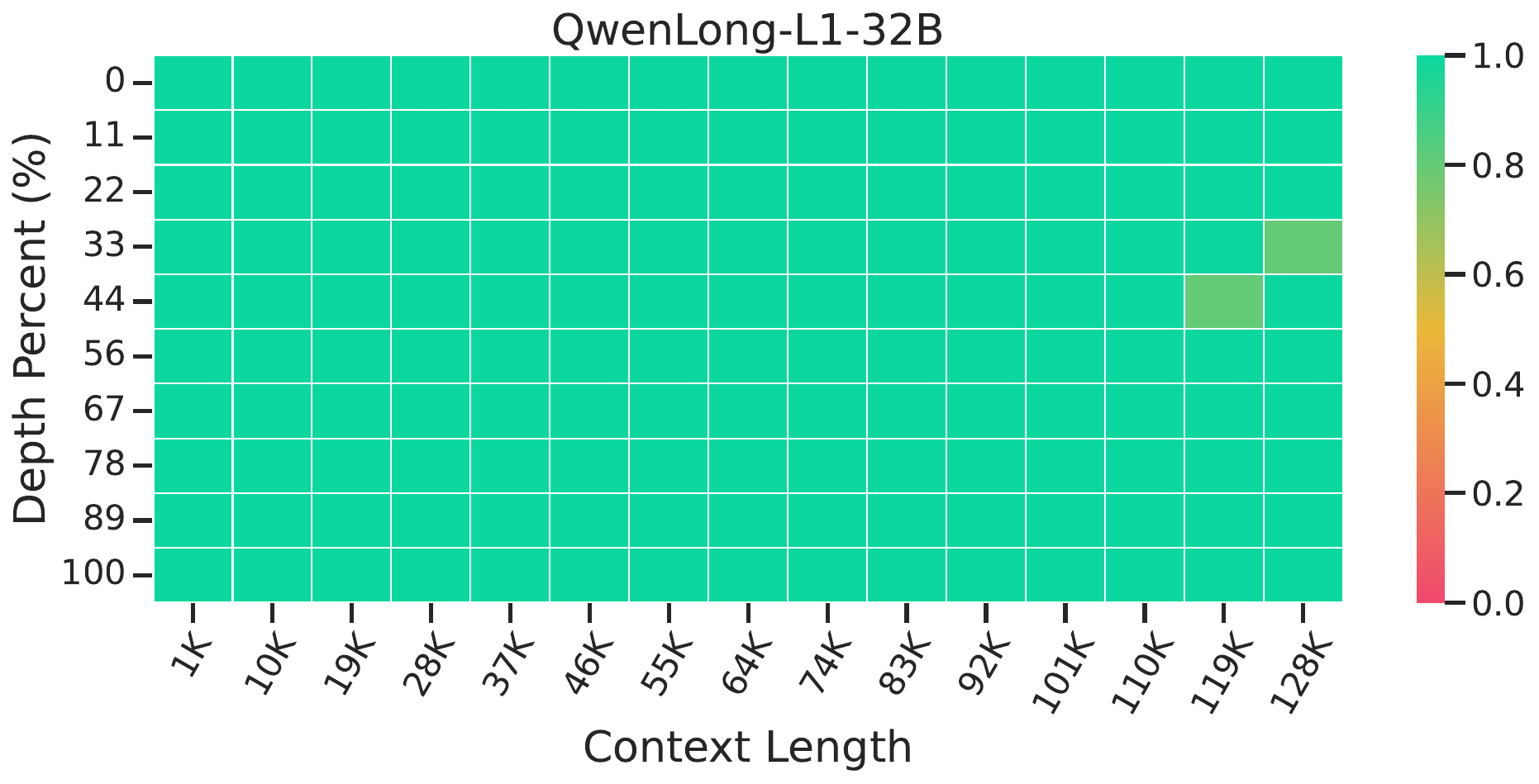}
	\end{subfigure}
	\caption{Needle in a Haystack retrieval across document depths. The base Qwen2.5-7B-Instruct does not fully pass the benchmark, whereas {\sysname}-7B achieves perfect 100\% retrieval accuracy.}
	\label{fig:niah1}
\end{figure}

\noindent\textbf{Improved long-context retrieval}. We evaluate the impact of different approaches on retrieval using the Needle in a Haystack benchmark, which measures a model’s ability to find “needles” from long documents at varying depths. As shown in Fig.~\ref{fig:niah1}, the base Qwen2.5-7B-Instruct fails to fully pass this benchmark. In contrast, our {\sysname} improves retrieval substantially, with {\sysname}-7B achieving perfect accuracy across all depths. Other approaches remain limited, with R1-Distill-7B unable to retrieve beyond 20K and even the larger QwenLong-L1-32B failing to achieve a full pass.

\subsection{Ablation Study}

\begin{figure}[t]
	\centering
	\includegraphics[width=1\linewidth]{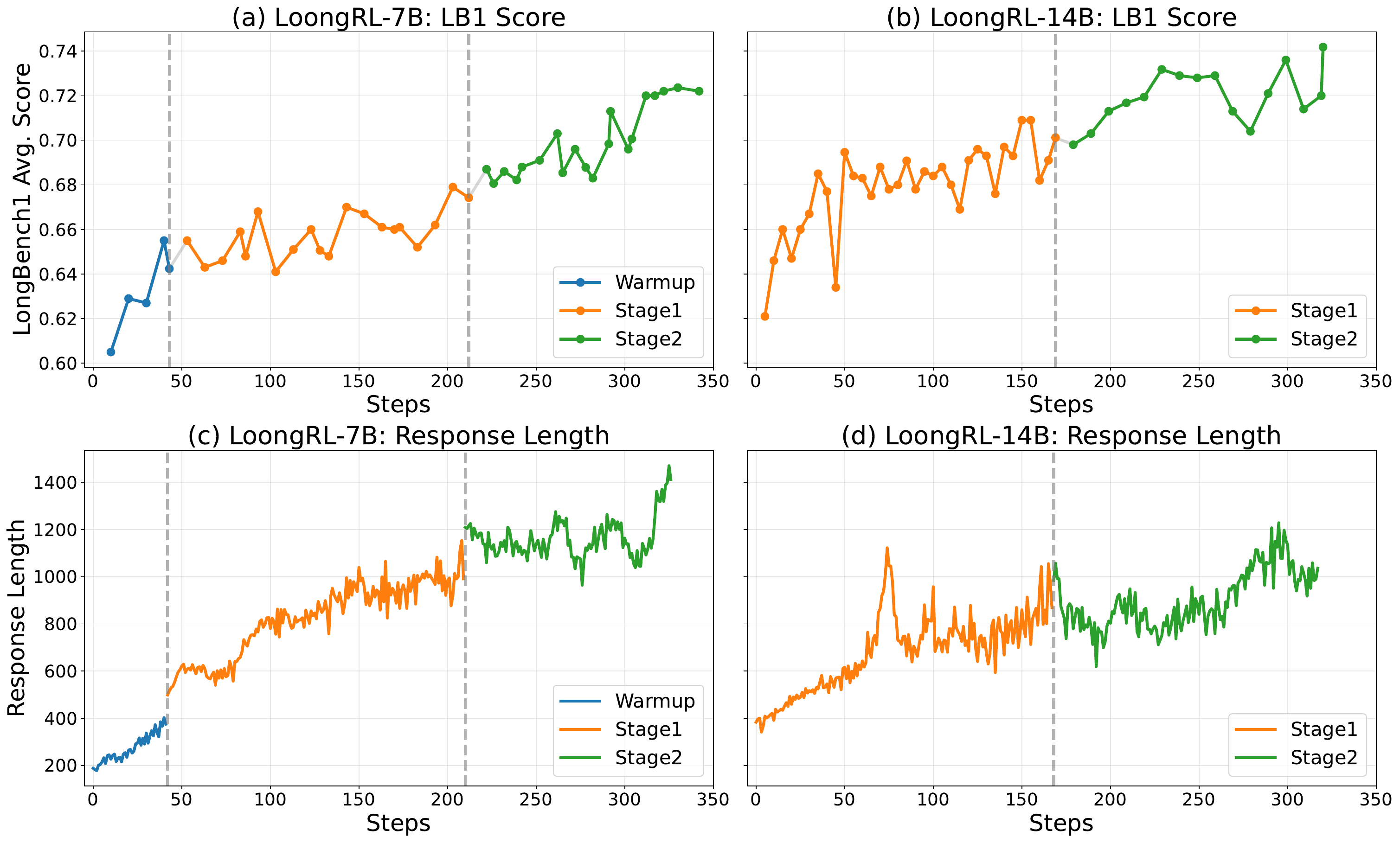}
	\caption{Long-context reasoning accuracy and  training response lengths throughout RL training. }
	\label{fig:rlstage}
		\vspace{-2ex}
\end{figure}

\noindent\textbf{Multi-stage RL training sustains improvements}.   To understand how {\sysname} achieves strong performance, we report step-by-step gains and average training lengths for 7B and 14B across the three RL stages. As shown in Fig.~\ref{fig:rlstage} (c,d), average response length steadily increases throughout training. Fig.~\ref{fig:rlstage} (a,b) presents long-context reasoning accuracy, which grows consistently across each stage, demonstrating the effectiveness of the multi-stage RL curriculum.

\begin{table}[t]
	\centering
	\caption{Ablation study on the effectiveness of KeyChain data.}
	\setlength{\tabcolsep}{5pt}
	\resizebox{0.98\textwidth}{!}{
	\begin{tabular}{lcccccg}
		\toprule
		{Models} & {HotpotQA} & {2WikiMultiHopQA} & { MuSiQue} & {NarrativeQA} & {QASPER}& Avg. \\
		\midrule
		Qwen2.5-7B-Instruct & 69.5& 50.5& 34.0& 44.5&46.0 &48.9\\
		{\sysname}-7B (no KeyChain Data)& 80.3 & 84.7 & 58.5 & 53.0 & 54.5 & 66.2 \\
	\bf{\sysname}-7B& \bf 83.1 & \bf 91.1 & \bf 65.6 & \bf 58.4 &\bf  63.6 &\bf  72.4 \\
		\bottomrule
	\end{tabular}}
	\label{tab:ablation-keychain1}
		\vspace{-2ex}
\end{table}

\noindent\textbf{Ablation on the KeyChain data}. Our KeyChain training data effectively encourages models to acquire new long-context reasoning patterns during RL. To evaluate its effectiveness, we replace it with an equal amount of regular long-context multi-hop QA data on Qwen2.5-7B-Instruct while keeping all other RL settings identical. Table~\ref{tab:ablation-keychain1} shows the comparison results.  RL with regular QA data yields moderate gains (66.2), whereas incorporating KeyChain data drives a substantial leap to 72.4, reaching frontier-level performance.

Moreover, as shown in Fig.~\ref{fig:reasoningpattern}(b), models trained with regular long-context multi-hop data exhibit a mixed reasoning-with-retrieval pattern. They often lack an explicit planning step and do not perform careful reason over the retrieved information, making them more prone to errors. This demonstrates that KeyChain not only significantly enhances long-context reasoning but also unlocks capabilities that cannot be achieved with conventional QA data, highlighting its unique and critical role.
\begin{table}[t]
	\setlength{\tabcolsep}{5pt}
	\centering
	\caption{Ablation study on the different answer verifiers on the 7B.}
	\resizebox{0.98\textwidth}{!}{	\begin{tabular}{@{\hskip0pt}l@{\hskip6pt}c@{\hskip6pt}c@{\hskip6pt}c@{\hskip6pt}c@{\hskip6pt}c@{\hskip6pt}g@{\hskip0pt}}
		\toprule
		 Reward Verifier & {HotpotQA} & {2WikiMultiHopQA} & { MuSiQue} & {NarrativeQA} & {QASPER}& Avg. \\
		\midrule
		F1 score          & 79.5 & 86.4 & 58.0 & 46.6 & 55.0 & 65.1  \\
		LLM-as-a-judge  & 80.0 & 87.6 & 60.0 & 52.3 & 54.5 & 65.2  \\
			Exact match  &  82.7 &	\bf 91.3 &	\bf 66.3 &	51.0 &	54.9  & 69.2     \\
		\bf Two-way Substring Exact Match (ours) & \bf 83.1 & 91.1 & 65.6 & \bf 58.4 & \bf 63.6 & \bf 72.4 \\
		\bottomrule
	\end{tabular}}
	\label{tbl:verifier}
	\vspace{-4ex}
\end{table}

\noindent\textbf{Ablation on the answer verifier}. To evaluate our two-way substring exact match for verifying answer correctness, we compare it with three widely used  baselines on Qwen2.5-7B-Instruct: (i) F1 score between extracted answers and the ground truth~\citep{shi2025search, chuang2025selfcite}; (ii) LLM-as-a-judge using DeepSeek-V3 to assess consistency with the ground truth; and (iii) exact match, requiring extract answer to match the ground truth exactly. As shown in Table~\ref{tbl:verifier}, F1 and LLM-as-a-judge yield moderate gains, while exact match performs better but is overly strict, penalizing essentially correct answers with minor formatting differences.  In contrast, our two-way substring exact match  maintains high precision while allowing variations, boosting long-context reasoning scores to 72.4 and clearly demonstrating its practical reliability  for RL training.

\section{Conclusion}
\vspace{-1ex}
This work introduces {\sysname}, a data-driven reinforcement learning approach for advanced long-context reasoning. By creating a novel dataset, KeyChain, which transforms standard multi-hop questions into high-difficulty tasks, {\sysname} trains models to develop a ``plan-retrieve-reason-recheck" thinking pattern. A key finding is that this emergent reasoning ability generalizes remarkably well. Models trained on 16K token contexts can effectively solve tasks up to 128K tokens. Our resulting  {\sysname}-14B model achieves a 74.2 score on long-context QA benchmarks, rivaling much larger frontier models like o3-mini and DeepSeek-R1. These significant gains are achieved while successfully preserving the model's crucial short-context reasoning and retrieval capabilities.


\section*{Reproducibility statement}
We have made extensive efforts to ensure the reproducibility of our work. Details of the GRPO algorithm and hyperparameters are provided in Section~\ref{sec:grpo} and Section~\ref{sec:setup}. We provide our training prompt template in Appendix~\ref{appendix:training_prompt}. The datasets used in our experiments are described in Table~\ref{tab:datasets}. To further facilitate reproducibility, the supplementary materials include  (i) our RL training code, (ii) the code for synthesizing KeyChain data, and (iii) several representative samples of the synthesized KeyChain data. These resources, together with the descriptions in the main text and appendix, provide all necessary information for replicating our results.




\bibliographystyle{plainnat}
\bibliography{references}

\begin{thebibliography}{41}
\providecommand{\natexlab}[1]{#1}
\providecommand{\url}[1]{\texttt{#1}}
\expandafter\ifx\csname urlstyle\endcsname\relax
  \providecommand{\doi}[1]{doi: #1}\else
  \providecommand{\doi}{doi: \begingroup \urlstyle{rm}\Url}\fi

\bibitem[Achiam et~al.(2023)Achiam, Adler, Agarwal, Ahmad, Akkaya, Aleman,
  Almeida, Altenschmidt, Altman, Anadkat, et~al.]{achiam2023gpt}
Josh Achiam, Steven Adler, Sandhini Agarwal, Lama Ahmad, Ilge Akkaya,
  Florencia~Leoni Aleman, Diogo Almeida, Janko Altenschmidt, Sam Altman,
  Shyamal Anadkat, et~al.
\newblock Gpt-4 technical report.
\newblock \emph{arXiv preprint arXiv:2303.08774}, 2023.

\bibitem[Ahmad et~al.(2025)Ahmad, Narenthiran, Majumdar, Ficek, Jain, Huang,
  Noroozi, and Ginsburg]{ahmad2025opencodereasoning}
Wasi~Uddin Ahmad, Sean Narenthiran, Somshubra Majumdar, Aleksander Ficek,
  Siddhartha Jain, Jocelyn Huang, Vahid Noroozi, and Boris Ginsburg.
\newblock Opencodereasoning: Advancing data distillation for competitive
  coding.
\newblock \emph{arXiv preprint arXiv:2504.01943}, 2025.

\bibitem[Bai et~al.(2024{\natexlab{a}})Bai, Lv, Zhang, Lyu, Tang, Huang, Du,
  Liu, Zeng, Hou, Dong, Tang, and Li]{longbench1}
Yushi Bai, Xin Lv, Jiajie Zhang, Hongchang Lyu, Jiankai Tang, Zhidian Huang,
  Zhengxiao Du, Xiao Liu, Aohan Zeng, Lei Hou, Yuxiao Dong, Jie Tang, and
  Juanzi Li.
\newblock {L}ong{B}ench: A bilingual, multitask benchmark for long context
  understanding.
\newblock In \emph{Proceedings of the 62nd Annual Meeting of the Association
  for Computational Linguistics (Volume 1: Long Papers)}, pages 3119--3137,
  Bangkok, Thailand, August 2024{\natexlab{a}}. Association for Computational
  Linguistics.
\newblock \doi{10.18653/v1/2024.acl-long.172}.
\newblock URL \url{https://aclanthology.org/2024.acl-long.172}.

\bibitem[Bai et~al.(2024{\natexlab{b}})Bai, Tu, Zhang, Peng, Wang, Lv, Cao, Xu,
  Hou, Dong, Tang, and Li]{longbench2}
Yushi Bai, Shangqing Tu, Jiajie Zhang, Hao Peng, Xiaozhi Wang, Xin Lv, Shulin
  Cao, Jiazheng Xu, Lei Hou, Yuxiao Dong, Jie Tang, and Juanzi Li.
\newblock Longbench v2: Towards deeper understanding and reasoning on realistic
  long-context multitasks.
\newblock \emph{arXiv preprint arXiv:2412.15204}, 2024{\natexlab{b}}.

\bibitem[Chuang et~al.(2025)Chuang, Cohen-Wang, Shen, Wu, Xu, Lin, Glass, Li,
  and Yih]{chuang2025selfcite}
Yung-Sung Chuang, Benjamin Cohen-Wang, Shannon~Zejiang Shen, Zhaofeng Wu,
  Hu~Xu, Xi~Victoria Lin, James Glass, Shang-Wen Li, and Wen-tau Yih.
\newblock Selfcite: Self-supervised alignment for context attribution in large
  language models.
\newblock \emph{arXiv preprint arXiv:2502.09604}, 2025.

\bibitem[Dasigi et~al.(2021)Dasigi, Lo, Beltagy, Cohan, Smith, and
  Gardner]{qasper}
Pradeep Dasigi, Kyle Lo, Iz~Beltagy, Arman Cohan, Noah~A. Smith, and Matt
  Gardner.
\newblock A dataset of information-seeking questions and answers anchored in
  research papers.
\newblock In Kristina Toutanova, Anna Rumshisky, Luke Zettlemoyer, Dilek
  Hakkani-Tur, Iz~Beltagy, Steven Bethard, Ryan Cotterell, Tanmoy Chakraborty,
  and Yichao Zhou, editors, \emph{Proceedings of the 2021 Conference of the
  North American Chapter of the Association for Computational Linguistics:
  Human Language Technologies}, pages 4599--4610, Online, June 2021.
  Association for Computational Linguistics.
\newblock \doi{10.18653/v1/2021.naacl-main.365}.
\newblock URL \url{https://aclanthology.org/2021.naacl-main.365/}.

\bibitem[Ding et~al.(2024)Ding, Zhang, Zhang, Xu, Shang, Xu, Yang, and
  Yang]{longrope}
Yiran Ding, Li~Lyna Zhang, Chengruidong Zhang, Yuanyuan Xu, Ning Shang, Jiahang
  Xu, Fan Yang, and Mao Yang.
\newblock Longrope: Extending llm context window beyond 2 million tokens, 2024.
\newblock URL \url{https://arxiv.org/abs/2402.13753}.

\bibitem[Gandhi et~al.(2025)Gandhi, Chakravarthy, Singh, Lile, and
  Goodman]{gandhi2025cognitive}
Kanishk Gandhi, Ayush Chakravarthy, Anikait Singh, Nathan Lile, and Noah~D
  Goodman.
\newblock Cognitive behaviors that enable self-improving reasoners, or, four
  habits of highly effective stars.
\newblock \emph{arXiv preprint arXiv:2503.01307}, 2025.

\bibitem[Guo et~al.(2025)Guo, Yang, Zhang, Song, Zhang, Xu, Zhu, Ma, Wang, Bi,
  et~al.]{deepseekr1}
Daya Guo, Dejian Yang, Haowei Zhang, Junxiao Song, Ruoyu Zhang, Runxin Xu,
  Qihao Zhu, Shirong Ma, Peiyi Wang, Xiao Bi, et~al.
\newblock Deepseek-r1: Incentivizing reasoning capability in llms via
  reinforcement learning.
\newblock \emph{arXiv preprint arXiv:2501.12948}, 2025.

\bibitem[Hendrycks et~al.(2020)Hendrycks, Burns, Basart, Zou, Mazeika, Song,
  and Steinhardt]{hendrycks2020mmlu}
Dan Hendrycks, Collin Burns, Steven Basart, Andy Zou, Mantas Mazeika, Dawn
  Song, and Jacob Steinhardt.
\newblock Measuring massive multitask language understanding.
\newblock \emph{arXiv preprint arXiv:2009.03300}, 2020.

\bibitem[Hendrycks et~al.(2021)Hendrycks, Burns, Kadavath, Arora, Basart, Tang,
  Song, and Steinhardt]{hendrycks2021math}
Dan Hendrycks, Collin Burns, Saurav Kadavath, Akul Arora, Steven Basart, Eric
  Tang, Dawn Song, and Jacob Steinhardt.
\newblock Measuring mathematical problem solving with the math dataset.
\newblock \emph{arXiv preprint arXiv:2103.03874}, 2021.

\bibitem[Ho et~al.(2020)Ho, Duong~Nguyen, Sugawara, and
  Aizawa]{2wikimultihopqa}
Xanh Ho, Anh-Khoa Duong~Nguyen, Saku Sugawara, and Akiko Aizawa.
\newblock Constructing a multi-hop {QA} dataset for comprehensive evaluation of
  reasoning steps.
\newblock In Donia Scott, Nuria Bel, and Chengqing Zong, editors,
  \emph{Proceedings of the 28th International Conference on Computational
  Linguistics}, pages 6609--6625, Barcelona, Spain (Online), December 2020.
  International Committee on Computational Linguistics.
\newblock \doi{10.18653/v1/2020.coling-main.580}.
\newblock URL \url{https://aclanthology.org/2020.coling-main.580/}.

\bibitem[Hsieh et~al.(2024)Hsieh, Sun, Kriman, Acharya, Rekesh, Jia, Zhang, and
  Ginsburg]{ruler}
Cheng-Ping Hsieh, Simeng Sun, Samuel Kriman, Shantanu Acharya, Dima Rekesh, Fei
  Jia, Yang Zhang, and Boris Ginsburg.
\newblock {RULER}: What{'}s the real context size of your long-context language
  models?
\newblock In \emph{Proceedings of the 1st Conference on Language Modeling
  (COLM)}, Philadelphia, PA, 2024.
\newblock URL \url{https://arxiv.org/abs/2404.06654}.

\bibitem[Jaech et~al.(2024)Jaech, Kalai, Lerer, Richardson, El-Kishky, Low,
  Helyar, Madry, Beutel, Carney, Iftimie, Karpenko, Passos, Neitz, Prokofiev,
  Wei, Tam, Bennett, Kumar, Saraiva, Vallone, Duberstein, Kondrich, Mishchenko,
  Applebaum, Jiang, Nair, Zoph, Ghorbani, Rossen, Sokolowsky, Barak, McGrew,
  Minaiev, Hao, Baker, Houghton, McKinzie, Eastman, Lugaresi, Bassin, Hudson,
  Li, de~Bourcy, Voss, Shen, Zhang, Koch, Orsinger, Hesse, Fischer, Chan,
  Roberts, Kappler, Levy, Selsam, Dohan, Farhi, Mely, Robinson, Tsipras, Li,
  Oprica, Freeman, Zhang, Wong, Proehl, Cheung, Mitchell, Wallace, Ritter,
  Mays, Wang, Such, Raso, Leoni, Tsimpourlas, Song, von Lohmann, Sulit, Salmon,
  Parascandolo, Chabot, Zhao, Brockman, Leclerc, Salman, Bao, Sheng, Andrin,
  Bagherinezhad, Ren, Lightman, Chung, Kivlichan, O'Connell, Osband, Gilaberte,
  and Akkaya]{o1}
Aaron Jaech, Adam Kalai, Adam Lerer, Adam Richardson, Ahmed El-Kishky, Aiden
  Low, Alec Helyar, Aleksander Madry, Alex Beutel, Alex Carney, Alex Iftimie,
  Alex Karpenko, Alex~Tachard Passos, Alexander Neitz, Alexander Prokofiev,
  Alexander Wei, Allison Tam, Ally Bennett, Ananya Kumar, Andre Saraiva, Andrea
  Vallone, Andrew Duberstein, Andrew Kondrich, Andrey Mishchenko, Andy
  Applebaum, Angela Jiang, Ashvin Nair, Barret Zoph, Behrooz Ghorbani, Ben
  Rossen, Benjamin Sokolowsky, Boaz Barak, Bob McGrew, Borys Minaiev, Botao
  Hao, Bowen Baker, Brandon Houghton, Brandon McKinzie, Brydon Eastman, Camillo
  Lugaresi, Cary Bassin, Cary Hudson, Chak~Ming Li, Charles de~Bourcy, Chelsea
  Voss, Chen Shen, Chong Zhang, Chris Koch, Chris Orsinger, Christopher Hesse,
  Claudia Fischer, Clive Chan, Dan Roberts, Daniel Kappler, Daniel Levy, Daniel
  Selsam, David Dohan, David Farhi, David Mely, David Robinson, Dimitris
  Tsipras, Doug Li, Dragos Oprica, Eben Freeman, Eddie Zhang, Edmund Wong,
  Elizabeth Proehl, Enoch Cheung, Eric Mitchell, Eric Wallace, Erik Ritter,
  Evan Mays, Fan Wang, Felipe~Petroski Such, Filippo Raso, Florencia Leoni,
  Foivos Tsimpourlas, Francis Song, Fred von Lohmann, Freddie Sulit, Geoff
  Salmon, Giambattista Parascandolo, Gildas Chabot, Grace Zhao, Greg Brockman,
  Guillaume Leclerc, Hadi Salman, Haiming Bao, Hao Sheng, Hart Andrin, Hessam
  Bagherinezhad, Hongyu Ren, Hunter Lightman, Hyung~Won Chung, Ian Kivlichan,
  Ian O'Connell, Ian Osband, Ignasi~Clavera Gilaberte, and Ilge Akkaya.
\newblock Openai o1 system card.
\newblock \emph{CoRR}, abs/2412.16720, 2024.
\newblock URL \url{https://doi.org/10.48550/arXiv.2412.16720}.

\bibitem[Kamradt(2023)]{needlehaystack}
Greg Kamradt.
\newblock Needle in a haystack - pressure testing llms, 2023.
\newblock URL \url{https://github.com/gkamradt/LLMTest_NeedleInAHaystack}.

\bibitem[{Kimi Team} et~al.(2025){Kimi Team}, Du, Gao, Xing, Jiang, Chen, Li,
  Xiao, Du, Liao, Tang, Wang, Zhang, Yuan, Lu, Tang, Sung, Wei, Lai, Guo, Zhu,
  Ding, Hu, Yang, Zhang, Yao, Zhao, Lu, Li, Yu, Gao, Zheng, Yuan, Chen, Guo,
  Su, Wang, Zhao, Zhang, Liu, Yan, Wu, Shi, Ye, Yu, Dong, Zhang, Ma, Pan, Gong,
  Liu, Ma, Wei, Cao, Huang, Jiang, Gao, Xiong, He, Huang, Xu, Wu, He, Wei, Jia,
  Wu, Xu, Zu, Zhou, Pan, Charles, Li, Hu, Liu, Chen, Wang, Liu, Qin, Liu, Yang,
  Bao, Du, Wu, Wang, Zhou, Wang, Li, Zhu, Zhang, Wang, Yang, Huang, Huang, Xu,
  Yang, and Lin]{kimi1.5}
{Kimi Team}, Angang Du, Bofei Gao, Bowei Xing, Changjiu Jiang, Cheng Chen,
  Cheng Li, Chenjun Xiao, Chenzhuang Du, Chonghua Liao, Chuning Tang, Congcong
  Wang, Dehao Zhang, Enming Yuan, Enzhe Lu, Fengxiang Tang, Flood Sung, Guangda
  Wei, Guokun Lai, Haiqing Guo, Han Zhu, Hao Ding, Hao Hu, Hao Yang, Hao Zhang,
  Haotian Yao, Haotian Zhao, Haoyu Lu, Haoze Li, Haozhen Yu, Hongcheng Gao,
  Huabin Zheng, Huan Yuan, Jia Chen, Jianhang Guo, Jianlin Su, Jianzhou Wang,
  Jie Zhao, Jin Zhang, Jingyuan Liu, Junjie Yan, Junyan Wu, Lidong Shi, Ling
  Ye, Longhui Yu, Mengnan Dong, Neo Zhang, Ningchen Ma, Qiwei Pan, Qucheng
  Gong, Shaowei Liu, Shengling Ma, Shupeng Wei, Sihan Cao, Siying Huang, Tao
  Jiang, Weihao Gao, Weimin Xiong, Weiran He, Weixiao Huang, Weixin Xu, Wenhao
  Wu, Wenyang He, Xianghui Wei, Xianqing Jia, Xingzhe Wu, Xinran Xu, Xinxing
  Zu, Xinyu Zhou, Xuehai Pan, Y.~Charles, Yang Li, Yangyang Hu, Yangyang Liu,
  Yanru Chen, Yejie Wang, Yibo Liu, Yidao Qin, Yifeng Liu, Ying Yang, Yiping
  Bao, Yulun Du, Yuxin Wu, Yuzhi Wang, Zaida Zhou, Zhaoji Wang, Zhaowei Li,
  Zhen Zhu, Zheng Zhang, Zhexu Wang, Zhilin Yang, Zhiqi Huang, Zihao Huang,
  Ziyao Xu, Zonghan Yang, and Zongyu Lin.
\newblock Kimi k1.5: Scaling reinforcement learning with llms, 2025.
\newblock URL \url{https://arxiv.org/abs/2501.12599}.

\bibitem[Ko{\v{c}}isk{\'y} et~al.(2018)Ko{\v{c}}isk{\'y}, Schwarz, Blunsom,
  Dyer, Hermann, Melis, and Grefenstette]{narrativeqa}
Tom{\'a}{\v{s}} Ko{\v{c}}isk{\'y}, Jonathan Schwarz, Phil Blunsom, Chris Dyer,
  Karl~Moritz Hermann, G{\'a}bor Melis, and Edward Grefenstette.
\newblock The {N}arrative{QA} reading comprehension challenge.
\newblock \emph{Transactions of the Association for Computational Linguistics},
  6:\penalty0 317--328, 2018.
\newblock \doi{10.1162/tacl_a_00023}.
\newblock URL \url{https://aclanthology.org/Q18-1023/}.

\bibitem[Lambert et~al.(2024)Lambert, Morrison, Pyatkin, Huang, Ivison,
  Brahman, Miranda, Liu, Dziri, Lyu, et~al.]{lambert2024tulu}
Nathan Lambert, Jacob Morrison, Valentina Pyatkin, Shengyi Huang, Hamish
  Ivison, Faeze Brahman, Lester James~V Miranda, Alisa Liu, Nouha Dziri, Shane
  Lyu, et~al.
\newblock Tulu 3: Pushing frontiers in open language model post-training.
\newblock \emph{arXiv preprint arXiv:2411.15124}, 2024.

\bibitem[Li et~al.(2025)Li, Gong, Yang, Shan, Liu, Zhu, Zhang, Guo, Chen, Li,
  et~al.]{li2025minimax}
Aonian Li, Bangwei Gong, Bo~Yang, Boji Shan, Chang Liu, Cheng Zhu, Chunhao
  Zhang, Congchao Guo, Da~Chen, Dong Li, et~al.
\newblock Minimax-01: Scaling foundation models with lightning attention.
\newblock \emph{arXiv preprint arXiv:2501.08313}, 2025.

\bibitem[Li et~al.(2024{\natexlab{a}})Li, Verga, Sen, Yang, Viswanathan, Lewis,
  Watanabe, and Su]{li-etal-2024-alr2}
Huayang Li, Pat Verga, Priyanka Sen, Bowen Yang, Vijay Viswanathan, Patrick
  Lewis, Taro Watanabe, and Yixuan Su.
\newblock {ALR}$^2$: A retrieve-then-reason framework for long-context question
  answering.
\newblock \emph{arXiv preprint arXiv:2410.03227}, 2024{\natexlab{a}}.

\bibitem[Li et~al.(2024{\natexlab{b}})Li, Yang, Cheng, Liu, Yu, Yang, and
  Lam]{li-etal-2024-large}
Siheng Li, Cheng Yang, Zesen Cheng, Lemao Liu, Mo~Yu, Yujiu Yang, and Wai Lam.
\newblock Large language models can self-improve in long-context reasoning.
\newblock \emph{arXiv preprint arXiv:2411.08147}, 2024{\natexlab{b}}.

\bibitem[Li et~al.(2024{\natexlab{c}})Li, Liang, Lyu, and
  Wang]{li-etal-2024-making}
Yanyang Li, Shuo Liang, Michael Lyu, and Liwei Wang.
\newblock Making long-context language models better multi-hop reasoners.
\newblock In \emph{Proceedings of the 62nd Annual Meeting of the Association
  for Computational Linguistics (Volume 1: Long Papers)}, pages 2462--2475,
  Bangkok, Thailand, August 2024{\natexlab{c}}. Association for Computational
  Linguistics.
\newblock \doi{10.18653/v1/2024.acl-long.135}.

\bibitem[Lightman et~al.(2023)Lightman, Kosaraju, Burda, Edwards, Baker, Lee,
  Leike, Schulman, Sutskever, and Cobbe]{math-500}
Hunter Lightman, Vineet Kosaraju, Yura Burda, Harri Edwards, Bowen Baker, Teddy
  Lee, Jan Leike, John Schulman, Ilya Sutskever, and Karl Cobbe.
\newblock Let's verify step by step.
\newblock \emph{arXiv preprint arXiv:2305.20050}, 2023.

\bibitem[Ling et~al.(2025)Ling, Liu, Yan, Yang, Lin, Fan, Shen, Du, and
  Chen]{ling2025longreason}
Zhan Ling, Kang Liu, Kai Yan, Yifan Yang, Weijian Lin, Ting-Han Fan, Lingfeng
  Shen, Zhengyin Du, and Jiecao Chen.
\newblock Longreason: A synthetic long-context reasoning benchmark via context
  expansion.
\newblock \emph{arXiv preprint arXiv:2501.15089}, 2025.

\bibitem[Liu et~al.(2024)Liu, Feng, Xue, Wang, Wu, Lu, Zhao, Deng, Zhang, Ruan,
  et~al.]{liu2024deepseek}
Aixin Liu, Bei Feng, Bing Xue, Bingxuan Wang, Bochao Wu, Chengda Lu, Chenggang
  Zhao, Chengqi Deng, Chenyu Zhang, Chong Ruan, et~al.
\newblock Deepseek-v3 technical report.
\newblock \emph{arXiv preprint arXiv:2412.19437}, 2024.

\bibitem[Liu et~al.(2025)Liu, Zhang, Zhu, Dong, Zhou, Shang, Yang, and
  Yang]{rstarcoder}
Yifei Liu, Li~Lyna Zhang, Yi~Zhu, Bingcheng Dong, Xudong Zhou, Ning Shang, Fan
  Yang, and Mao Yang.
\newblock rstar-coder: Scaling competitive code reasoning with a large-scale
  verified dataset.
\newblock \emph{arXiv preprint arXiv:2505.21297}, 2025.

\bibitem[Peng et~al.(2023)Peng, Quesnelle, Fan, and Shippole]{peng2023yarn}
Bowen Peng, Jeffrey Quesnelle, Honglu Fan, and Enrico Shippole.
\newblock Yarn: Efficient context window extension of large language models.
\newblock \emph{arXiv preprint arXiv:2309.00071}, 2023.

\bibitem[{Qwen Team}(2024)]{qwen2.5}
{Qwen Team}.
\newblock Qwen2.5: A party of foundation models, September 2024.
\newblock URL \url{https://qwenlm.github.io/blog/qwen2.5/}.

\bibitem[Rajpurkar et~al.(2016)Rajpurkar, Zhang, Lopyrev, and
  Liang]{rajpurkar-etal-2016-squad}
Pranav Rajpurkar, Jian Zhang, Konstantin Lopyrev, and Percy Liang.
\newblock {SQ}u{AD}: 100{,}000+ questions for machine comprehension of text.
\newblock In \emph{Proceedings of the 2016 Conference on Empirical Methods in
  Natural Language Processing}, pages 2383--2392, Austin, Texas, November 2016.
  Association for Computational Linguistics.
\newblock \doi{10.18653/v1/D16-1264}.

\bibitem[Shang et~al.(2025{\natexlab{a}})Shang, Liu, Zhu, Zhang, Xu, Guan,
  Zhang, Dong, Zhou, Zhang, Xin, Miao, Li, Yang, and Yang]{rstar2agent}
Ning Shang, Yifei Liu, Yi~Zhu, Li~Lyna Zhang, Weijiang Xu, Xinyu Guan, Buze
  Zhang, Bingcheng Dong, Xudong Zhou, Bowen Zhang, Ying Xin, Ziming Miao,
  Scarlett Li, Fan Yang, and Mao Yang.
\newblock rstar2-agent: Agentic reasoning technical report, 2025{\natexlab{a}}.
\newblock URL \url{https://arxiv.org/abs/2508.20722}.

\bibitem[Shang et~al.(2025{\natexlab{b}})Shang, Zhang, Wang, Zhang, Lopez,
  Yang, Chen, and Yang]{longrope2}
Ning Shang, Li~Lyna Zhang, Siyuan Wang, Gaokai Zhang, Gilsinia Lopez, Fan Yang,
  Weizhu Chen, and Mao Yang.
\newblock Longrope2: Near-lossless llm context window scaling,
  2025{\natexlab{b}}.
\newblock URL \url{https://arxiv.org/abs/2502.20082}.

\bibitem[Shi et~al.(2025)Shi, Li, Wu, Liu, Fang, Cai, Zhang, and
  Wang]{shi2025search}
Yaorui Shi, Sihang Li, Chang Wu, Zhiyuan Liu, Junfeng Fang, Hengxing Cai,
  An~Zhang, and Xiang Wang.
\newblock Search and refine during think: Autonomous retrieval-augmented
  reasoning of llms.
\newblock \emph{arXiv preprint arXiv:2505.11277}, 2025.

\bibitem[Trivedi et~al.(2022)Trivedi, Balasubramanian, Khot, and
  Sabharwal]{musique}
Harsh Trivedi, Niranjan Balasubramanian, Tushar Khot, and Ashish Sabharwal.
\newblock Musique: Multihop questions via single-hop question composition,
  2022.
\newblock URL \url{https://arxiv.org/abs/2108.00573}.

\bibitem[Wan et~al.(2025)Wan, Shen, Liao, Shi, Li, Yang, Zhang, Huang, Zhou,
  and Yan]{wan2025qwenlong}
Fanqi Wan, Weizhou Shen, Shengyi Liao, Yingcheng Shi, Chenliang Li, Ziyi Yang,
  Ji~Zhang, Fei Huang, Jingren Zhou, and Ming Yan.
\newblock Qwenlong-l1: Towards long-context large reasoning models with
  reinforcement learning, 2025.
\newblock URL \url{https://arxiv.org/abs/2505.17667}.

\bibitem[Wang et~al.(2025{\natexlab{a}})Wang, Yu, Gao, Zheng, Liu, Lu, Dang,
  Chen, Yang, Zhang, Liu, Yang, Zhao, Yue, Song, Yu, Huang, and
  Lin]{wang2025beyond}
Shenzhi Wang, Le~Yu, Chang Gao, Chujie Zheng, Shixuan Liu, Rui Lu, Kai Dang,
  Xionghui Chen, Jianxin Yang, Zhenru Zhang, Yuqiong Liu, An~Yang, Andrew Zhao,
  Yang Yue, Shiji Song, Bowen Yu, Gao Huang, and Junyang Lin.
\newblock Beyond the 80/20 rule: High-entropy minority tokens drive effective
  reinforcement learning for llm reasoning, 2025{\natexlab{a}}.
\newblock URL \url{https://arxiv.org/abs/2506.01939}.

\bibitem[Wang et~al.(2025{\natexlab{b}})Wang, Yang, Zeng, Ren, Liu, Peng,
  Cheng, He, Wang, Gao, Chen, Wang, Du, and Shen]{wang2025reinforcement}
Yiping Wang, Qing Yang, Zhiyuan Zeng, Liliang Ren, Lucas Liu, Baolin Peng, Hao
  Cheng, Xuehai He, Kuan Wang, Jianfeng Gao, Weizhu Chen, Shuohang Wang,
  Simon~Shaolei Du, and Yelong Shen.
\newblock Reinforcement learning for reasoning in large language models with
  one training example.
\newblock \emph{arXiv preprint arXiv:2504.20571}, 2025{\natexlab{b}}.
\newblock URL \url{https://arxiv.org/abs/2504.20571}.

\bibitem[Yang et~al.(2025)Yang, Li, Yang, Zhang, Hui, Zheng, Yu, Gao, Huang,
  Lv, et~al.]{qwen3}
An~Yang, Anfeng Li, Baosong Yang, Beichen Zhang, Binyuan Hui, Bo~Zheng, Bowen
  Yu, Chang Gao, Chengen Huang, Chenxu Lv, et~al.
\newblock Qwen3 technical report.
\newblock \emph{arXiv preprint arXiv:2505.09388}, 2025.

\bibitem[Yang et~al.(2018)Yang, Qi, Zhang, Bengio, Cohen, Salakhutdinov, and
  Manning]{hotpotqa}
Zhilin Yang, Peng Qi, Saizheng Zhang, Yoshua Bengio, William Cohen, Ruslan
  Salakhutdinov, and Christopher~D. Manning.
\newblock {H}otpot{QA}: A dataset for diverse, explainable multi-hop question
  answering.
\newblock In Ellen Riloff, David Chiang, Julia Hockenmaier, and Jun{'}ichi
  Tsujii, editors, \emph{Proceedings of the 2018 Conference on Empirical
  Methods in Natural Language Processing}, pages 2369--2380, Brussels, Belgium,
  October-November 2018. Association for Computational Linguistics.
\newblock \doi{10.18653/v1/D18-1259}.
\newblock URL \url{https://aclanthology.org/D18-1259/}.

\bibitem[Yen et~al.(2024)Yen, Gao, Hou, Ding, Fleischer, Izsak, Wasserblat, and
  Chen]{yen2024helmet}
Howard Yen, Tianyu Gao, Minmin Hou, Ke~Ding, Daniel Fleischer, Peter Izsak,
  Moshe Wasserblat, and Danqi Chen.
\newblock Helmet: How to evaluate long-context language models effectively and
  thoroughly.
\newblock \emph{arXiv preprint arXiv:2410.02694}, 2024.

\bibitem[Yu et~al.(2025)Yu, Zhang, Zhu, Yuan, Zuo, Yue, Dai, Fan, Liu, Liu,
  Liu, Lin, Lin, Ma, Sheng, Tong, Zhang, Zhang, Zhang, Zhu, Zhu, Chen, Chen,
  Wang, Yu, Song, Wei, Zhou, Liu, Ma, Zhang, Yan, Qiao, Wu, and
  Wang]{dapo_math}
Qiying Yu, Zheng Zhang, Ruofei Zhu, Yufeng Yuan, Xiaochen Zuo, Yu~Yue, Weinan
  Dai, Tiantian Fan, Gaohong Liu, Lingjun Liu, Xin Liu, Haibin Lin, Zhiqi Lin,
  Bole Ma, Guangming Sheng, Yuxuan Tong, Chi Zhang, Mofan Zhang, Wang Zhang,
  Hang Zhu, Jinhua Zhu, Jiaze Chen, Jiangjie Chen, Chengyi Wang, Hongli Yu,
  Yuxuan Song, Xiangpeng Wei, Hao Zhou, Jingjing Liu, Wei-Ying Ma, Ya-Qin
  Zhang, Lin Yan, Mu~Qiao, Yonghui Wu, and Mingxuan Wang.
\newblock Dapo: An open-source llm reinforcement learning system at scale,
  2025.
\newblock URL \url{https://arxiv.org/abs/2503.14476}.

\bibitem[Zhou et~al.(2023)Zhou, Lu, Mishra, Brahma, Basu, Luan, Zhou, and
  Hou]{zhou2023instruction}
Jeffrey Zhou, Tianjian Lu, Swaroop Mishra, Siddhartha Brahma, Sujoy Basu,
  Yi~Luan, Denny Zhou, and Le~Hou.
\newblock Instruction-following evaluation for large language models.
\newblock \emph{arXiv preprint arXiv:2311.07911}, 2023.

\end{thebibliography}

\appendix

\section{Appendix}
\subsection{Use of Large Language Models in Paper Writing}
In this work, we used large language models (LLMs) solely as general-purpose tools. Specifically, we employed LLMs to improve the clarity and readability of the paper. Additionally, during our ablation experiments, we evaluated the effectiveness of the answer verifier by using DeepSeek-V3 as the baseline in an LLM-as-a-judge setting.
\newpage
\subsection{Training Prompt Template}
\label{appendix:training_prompt}
For reproducibility, we include the exact prompt format used during training (see Figure~\ref{fig:train_prompt}). 
The model was trained to first generate intermediate reasoning enclosed in \verb|<think> ... </think>|, and then provide the final answer enclosed in \verb|\boxed{}|.

\begin{figure}[h]
\begin{tcolorbox}[colback=white,
	colframe=gray!75!black, title=System Prompt]
\small
A conversation between User and Assistant. The User asks a question, and the Assistant solves it. The Assistant first thinks about the reasoning process in the mind and then provides the User with the answer. The reasoning process is enclosed within \texttt{<think>} \texttt{</think>} and answer is enclosed within \verb|\boxed{}| tags, respectively, i.e., \texttt{<think>} reasoning process here \texttt{</think>} \verb|\boxed{answer here}|.
\end{tcolorbox}

\caption{System prompt used during training}
\label{fig:train_prompt}
\end{figure}

\subsection{Example of KeyChain-augmented Training Data}
\label{appendix:training_data_example}

\begin{figure}[ht]
	\centering
	\includegraphics[width=1\textwidth]{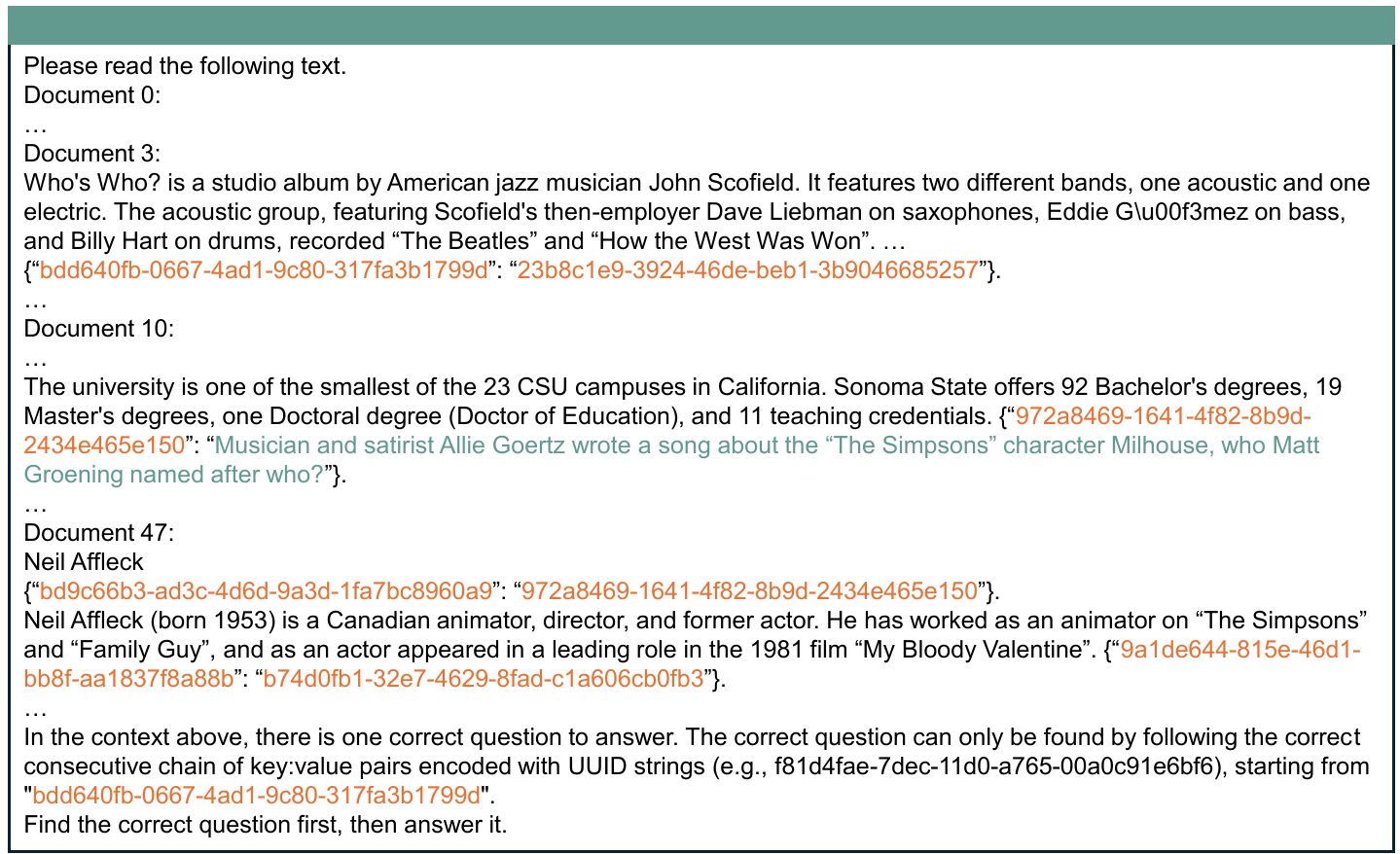}
	\caption{A skeleton of KeyChain-augmented training data used in LoongRL.}
	\label{fig:question}
\end{figure}

Figure~\ref{fig:question} shows a sample skeleton of our training data, illustrating how we carry out the Key-Chain augmentation for long-context reinforcement learning.

\newpage
\subsection{Comparison between traces of KeyChain vs.\ Non-KeyChain models}

Here we contrast two models trained with same hyperparameters but different data. The KeyChain-augmented model shows structured, step-by-step reasoning (colored as green) and clear re-checking when ambiguity arises, while the non-KeyChain model often shortcuts retrieval and proceeds to reasoning prematurely, leading to errors (colored as red). The following traces illustrate these differences on the same question: 

\textbf{Question:} Gary L. Bennett was a part of the space missions that have a primary destination of what celestial body?
\begin{center}
	\fontsize{8.2}{8.2} \selectfont
\begin{tcolorbox}[colback=white,
	colframe=gray!75!black, title={KeyChain-trained Model}]
\textcolor{ForestGreen}{To determine the primary destination of the space missions that Gary L. Bennett was a part of, we need to identify the space missions he worked on from the given passages and then find the primary destination of those missions.}

From Passage 1, we can see that Gary L. Bennett worked on the following space missions:
"His professional career has included work on the Voyager, Galileo, and Ulysses space missions,"

\textcolor{ForestGreen}{Now, we need to find the primary destination of these missions. Let's check Passage 5 for information about the Ulysses mission:}

"Ulysses ( yoo-LISS-eez, UK also YOO-liss-eez) was a robotic space probe whose primary mission was to orbit the Sun and study it at all latitudes."

Therefore, the primary destination of the Ulysses space mission is the Sun.

\textcolor{ForestGreen}{Although we should check the other missions mentioned (Voyager and Galileo) for completeness, we can see from Passage 3:}

"No missions were planned to coincide with the transit of Earth visible from Mars on 11 May 1984 and the Viking missions had been terminated a year previously. Consequently, the next opportunity to observe such an alignment will be in 2084."
"On 21 December 2012, the Cassini-Huygens probe, in orbit around Saturn, observed the planet Venus transiting the Sun."
"These are Schiaparelli EDM lander, the Mars Polar Lander, and Deep Space 2. These were all sterilized for surface missions but not for special regions (Viking pre-sterilization only). Mars Polar Lander, and Deep Space 2 crashed into the polar regions which are now treated as special regions because of the possibility of forming liquid brines."

And from Passage 4:
"Ulysses (spacecraft)"
"The spacecraft was originally named Odysseus, because of its lengthy and indirect trajectory to study the solar poles. It was renamed Ulysses, the Latin translation of "Odysseus", at ESA's request in honor not only of Homer's mythological hero but also of Dante's character in the Inferno. Ulysses was originally scheduled for launch in May 1986 aboard the Space Shuttle Challenger on STS-61-F. Due to the 28 January 1986 loss of Challenger, the launch of Ulysses was delayed until 6 October 1990 aboard Discovery (mission STS-41)."

The Voyager mission is known to have a primary destination of the outer planets, specifically Jupiter, Saturn, Uranus, and Neptune. The Galileo mission had a primary destination of Jupiter.

\textcolor{ForestGreen}{However, the question specifically asks about the space missions that Gary L. Bennett was a part of, and from the information provided, the Ulysses mission is clearly stated to have the Sun as its primary destination.}

Therefore, the primary destination of the space missions that Gary L. Bennett was a part of, specifically the Ulysses mission, is the Sun.

\(\boxed{\text{\textcolor{ForestGreen}{the Sun}}}\)
\end{tcolorbox}
\end{center}
\begin{center}
	\fontsize{8.2}{8.2} \selectfont
\begin{tcolorbox}[colback=white,
	colframe=gray!75!black, title={No KeyChain Model}]
\textcolor{red}{(Lack of systematic decomposition of the question)}
\\
First, we need to identify the space missions mentioned in the passage that Gary L. Bennett was involved with. From the biography provided, we can see that Bennett was involved in several space missions, including the Voyager, Galileo, and Ulysses space missions. The question asks for the primary destination of these missions.

- For the Voyager mission, the passage states: "He was the flight safety manager for the radioisotope power sources currently in use on the Voyager 1 and Voyager 2 spacecraft (which went to Jupiter, Saturn, Uranus, Neptune and beyond)."\\
- For the Galileo mission, it mentions: "Prior to coming to NASA, Bennett held key positions in DoE's space radioisotope power program, including serving as Director of Safety and Nuclear Operations for the radioisotope power sources that were used on the Galileo mission to Jupiter and that are being used on the Ulysses mission to explore the polar regions of the Sun."\\
- For the Ulysses mission, the passage states: "Ulysses ( yoo-LISS-eez, UK also  YOO-liss-eez) was a robotic space probe whose primary mission was to orbit the Sun and study it at all latitudes."

\textcolor{red}{From these statements, we can deduce that Bennett was involved in missions to Jupiter, Saturn, Uranus, Neptune, and the Sun.} However, the primary destinations mentioned directly in relation to Bennett's roles are Jupiter for the Galileo and Ulysses missions, and multiple planets including Jupiter, Saturn, Uranus, and Neptune for the Voyager missions.

\textcolor{red}{Given the specific mention of Bennett's role in the Galileo mission to Jupiter, and the broader statement about the Ulysses mission studying the Sun, Jupiter stands out as a primary destination that Bennett was directly involved with.}

\(\boxed{\text{\textcolor{red}{Jupiter}}}\)
\end{tcolorbox}
\end{center}

\subsection{Trajectory Examples on Long-context Retrieval}
\label{sec:appretrieval}
We show trajectories to highlight the impact  of KeyChain data on general long-context retrieval. In the Ruler VT examples in~\ref{fig:traj_keychain}, the KeyChain-trained model demonstrates clearer, stepwise retrieval, while the non-KeyChain-trained model conflates intermediate variable assignments and propagates errors, showing KeyChain data's role in eliciting structured and correct retrieval behaviors.

\begin{center}
	\fontsize{8.5}{8.5} \selectfont
\begin{tcolorbox}[colback=white,
	colframe=gray!75!black, title=RL Trajectory Example with KeyChain Data: Long-Context Retrieval on Ruler VT]

	$<$think$>$ 
	However, the specific value 92018 is only directly assigned to one variable in the text provided. Let's break it down step by step:
	
	1. The first assignment in the text is:
	“VAR QPE = 64886 The grass is green. The sky is blue. The sun is yellow. Here we go. There and back again.”
	This assigns the value 64886 to the variable QPE.
	
	2. The next assignments form a chain:
	“VAR SEJ = VAR QPE  The grass is green. The sky is blue. The sun is yellow. Here we go. There and back again.”
	This assigns the value of QPE (which is 64886) to the variable SEJ.
	
	3. “VAR ZQO = VAR SEJ  The grass is green. The sky is blue. The sun is yellow. Here we go. There and back again.”
	This assigns the value of SEJ (which is 64886) to the variable ZQO.
	
	4. “VAR RVU = VAR ZQO  The grass is green. The sky is blue. The sun is yellow. Here we go. There and back again.”
	This assigns the value of ZQO (which is 64886) to the variable RVU.
	
	5. “VAR FAI = VAR RVU  The grass is green. The sky is blue. The sun is yellow. Here we go. There and back again.”
	This assigns the value of RVU (which is 64886) to the variable FAI.
	
	6. The value 92018 is assigned to the variable SGMLJ:
	“\textcolor{orange}{VAR SGMLJ = 92018} The grass is green. The sky is blue. The sun is yellow. Here we go. There and back again.”
	
	7. Then, it forms a new chain:
	“\textcolor{orange}{VAR PBDME = VAR SGMLJ}  The grass is green. The sky is blue. The sun is yellow. Here we go. There and back again.”
	This assigns the value of SGMLJ (which is 92018) to the variable PBDME.
	
	8. “\textcolor{orange}{VAR EANSM = VAR PBDME}  The grass is green. The sky is blue. The sun is yellow. Here we go. There and back again.”
	This assigns the value of PBDME (which is 92018) to the variable EANSM.
	
	9. “\textcolor{orange}{VAR QPKBX = VAR EANSM}  The grass is green. The sky is blue. The sun is yellow. Here we go. There and back again.”
	This assigns the value of EANSM (which is 92018) to the variable QPKBX.
	
	10. “\textcolor{orange}{VAR YYZJM = VAR EANSM}  The grass is green. The sky is blue. The sun is yellow. Here we go. There and back again.”
	This assigns the value of EANSM (which is 92018) to the variable YYZJM.
	
	So, the variables that are assigned the value 92018 are:
	SGMLJ, PBDME, EANSM, QPKBX, and YYZJM. $<$/think$>$ 
	
	$<$think$>$ The variables that are assigned the value 92018 are:
	
	1. VAR SGMLJ = 92018
	2. VAR PBDME = VAR SGMLJ
	3. VAR EANSM = VAR PBDME
	4. VAR QPKBX = VAR EANSM5. VAR YYZJM = VAR EANSM
	Thus, the variables are SGMLJ, PBDME, EANSM, QPKBX, and YYZJM.  $<$/think$>$ 
	
	\verb|\|boxed\{SGMLJ, PBDME, EANSM, QPKBX, YYZJM\}
	\label{fig:traj_keychain}
\end{tcolorbox}
\end{center}

\begin{tcolorbox}[colback=white,
	colframe=gray!75!black, title=RL Trajectory Example WITHOUT KeyChain Data: Long-Context Retrieval on Ruler VT]

	$<$think$>$ 
	The reasoning process involves identifying the lines where the value 92018 is first assigned to a variable, which is  “\textcolor{orange}{VAR PBDME = 92018}”. Then, tracking this value through all subsequent assignments where this variable is reassigned to other variables. The variables that inherit this value through the assignments are \textcolor{red}{EANSM, YYZJM, SGMLJ, and FAI}. Since the question asks for variables assigned the value 92018, and FAI is assigned 92018 through the chain \textcolor{red}{VAR PBDME = VAR SGMLJ, VAR SGMLJ = 92018, and so on}, the final list of variables is PBDME, EANSM, YYZJM, SGMLJ, and FAI. $<$/think$>$
	
	\verb|\|boxed\{PBDME, EANSM, YYZJM, SGMLJ, FAI\}
	\label{fig:traj_nokeychain}
\end{tcolorbox}

\subsection{{\sysname} Training Dynamics}

\begin{figure}[t!]
	\centering
	\includegraphics[width=\textwidth]{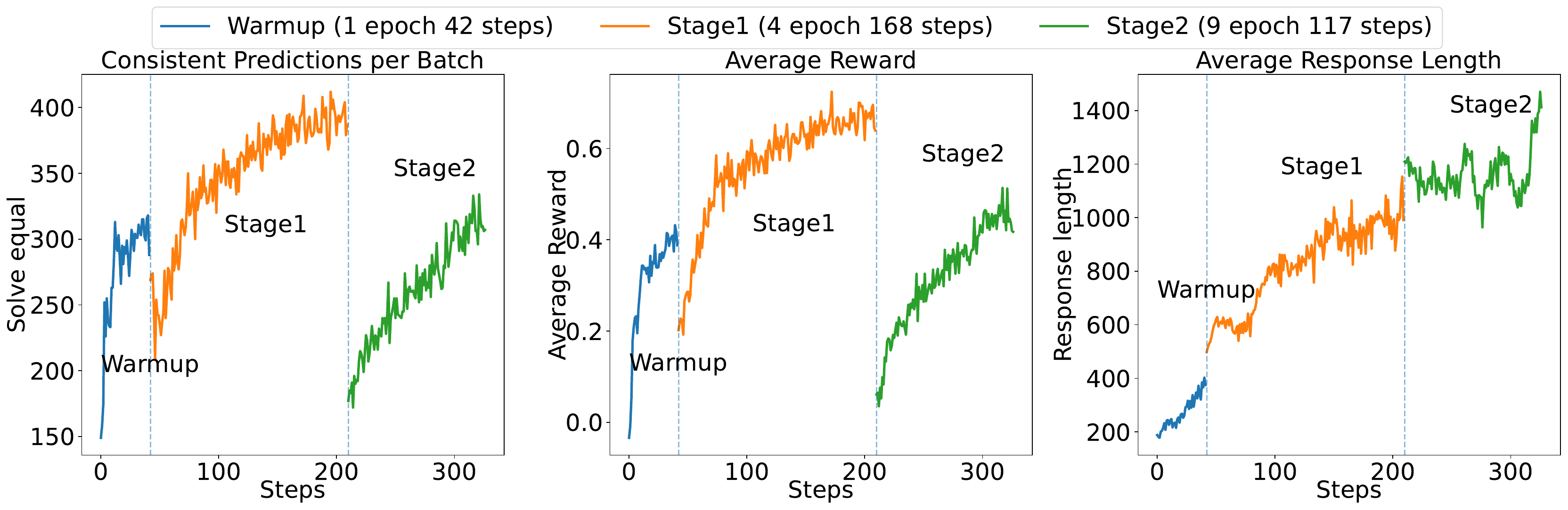}
	\caption{Training metrics for the three-stage schedule. Vertical dashed lines mark the transitions \emph{Warmup}\,$\rightarrow$\,\emph{Stage II (distractor-augmented)} and \emph{Stage II}\,$\rightarrow$\,\emph{Stage III (hard-mined)}; stage lengths correspond to our setup (about 42, 168, and 117 steps).}
	\label{fig:train}
\end{figure}

\begin{figure}[t!]
	\centering
	\includegraphics[width=\textwidth]{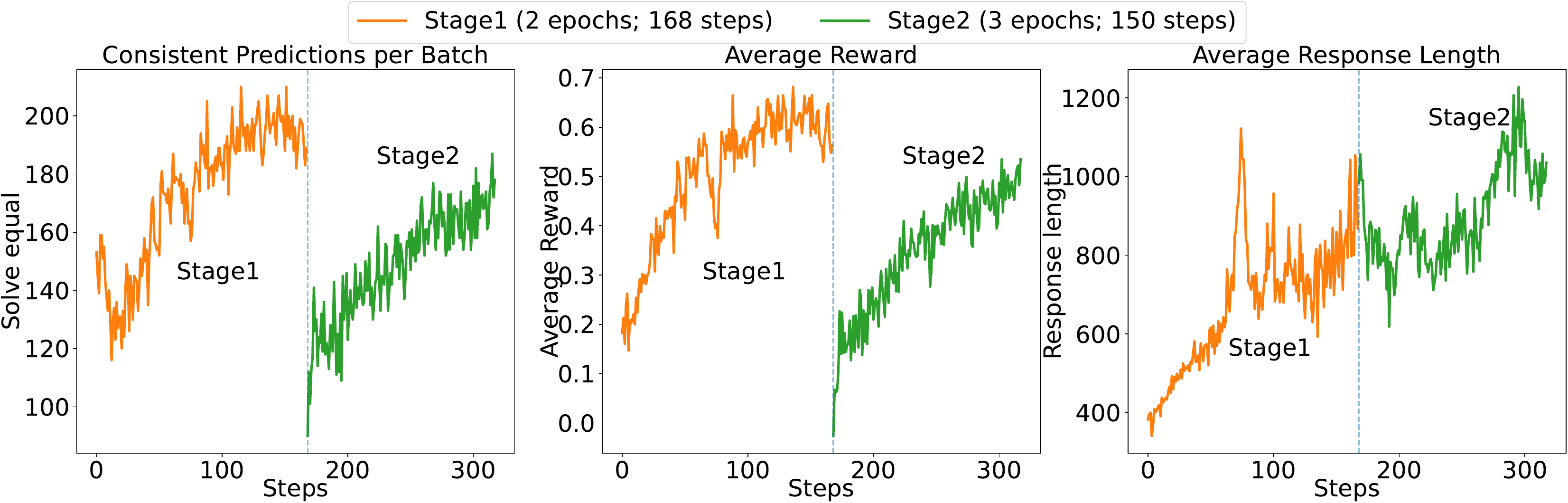}
	\caption{Training metrics for the LoongRL-14B's two-stage schedule. Vertical dashed lines mark the transitions \emph{Stage I (distractor-augmented)} and $\rightarrow$\,\emph{Stage III (hard-mined)}; stage lengths correspond to our setup (about 168, and 150 steps).}
	\label{fig:train14b}
\end{figure}
Fig.~\ref{fig:train} and Fig.~\ref{fig:train14b} show the training dynamics for the 7B and 14B models, respectively. Unlike the 7B setting, the 14B model is trained using only two curriculum stages, \textit{distractor} and \textit{hard-mined}, skipping the initial warm-up due to its strong base capabilities. As shown in two figures, a similar cyclical trend emerges: counts of consistently correct predictions and average rewards rise within each stage, reset when transitioning to a harder task pool, and increase again. This multi-stage training continues to provide informative learning signals and prevents saturation, while the steady growth in response length indicates that the model persistently extends its retrieval and reasoning chains in long-context multi-hop reasoning.

\subsection{LongBench-v2}

\begin{table*}[htbp!]
\centering
\renewcommand{\arraystretch}{0.9}
\setlength{\tabcolsep}{3pt}
\caption{Comparison of LoongRL models with other baselines on the LongBench-v2 benchmark, grouped by Difficulty, Length, and Task Type.}
\resizebox{\textwidth}{!}{%
\begin{tabular}{@{\hskip0pt}l@{\hskip6pt}g@{\hskip6pt}c@{\hskip6pt}c@{\hskip6pt}c@{\hskip6pt}c@{\hskip6pt}c@{\hskip6pt}c@{\hskip6pt}c@{\hskip6pt}c@{\hskip6pt}c@{\hskip6pt}c@{\hskip6pt}c@{\hskip0pt}}
\toprule
\textbf{Model} 
& \textbf{Overall} 
& \multicolumn{2}{c}{Difficulty} 
& \multicolumn{3}{c}{Length} 
& \multicolumn{6}{c}{Task Type} \\
\cmidrule(lr){2-2}
\cmidrule(lr){3-4}
\cmidrule(lr){5-7}
\cmidrule(lr){8-13}
&  & \textbf{Easy} & \textbf{Hard} 
& \textbf{Short} & \textbf{Medium} & \textbf{Long} 
& \textbf{Long ICL} & \textbf{Long SDU} & \textbf{Code} 
& \textbf{SingleDoc QA} & \textbf{Long Dialogue} & \textbf{MultiDoc QA} \\
\midrule
o3-mini & 46.4 & 52.9 & 42.4 & \underline{\textbf{56.1}} & 41.2 & 40.2 & 43.2 & 40.6 & 46.0 & 46.8 & \underline{\textbf{71.8}} & 41.6 \\
GPT-4o & 48.3 & \underline{\textbf{61.8}} & 40.3 & 46.2 & 48.4 & \underline{\textbf{51.0}} & \underline{\textbf{57.6}} & 44.4 & \underline{\textbf{66.7}} & 46.2 & 50.0 & 40.8 \\
QwQ-32B & \underline{\textbf{51.2}} & 57.8 & \underline{\textbf{47.1}} & \underline{\textbf{53.7}} & \underline{\textbf{51.2}} & 46.5 & 54.6 & 35.6 & 50.4 & \underline{\textbf{51.8}} & 56.9 & \underline{\textbf{49.9}} \\
R1-Distill-LLaMa-70B & 34.2 & 35.4 & 33.4 & 47.2 & 28.4 & 24.1 & 28.4 & 15.2 & 32.0 & 37.7 & 46.2 & 35.2 \\
\midrule
Qwen2.5-7B-Instruct & 31.2 & 32.3 & 30.5 & \textbf{42.8} & 24.7 & 25.0 & 25.9 & 30.3 & 42.0 & 35.4 & 35.9 & 23.2 \\
R1-Distill-Qwen-7B & 27.0 & 29.2 & 25.7 & 30.6 & 23.7 & 27.8 & 21.0 & 18.2 & 32.0 & 25.1 & 33.3 & \textbf{32.0} \\
\textbf{LoongRL-7B} & \textbf{36.2} & \textbf{41.1} & \textbf{33.1} & 40.6 & \textbf{34.4} & \textbf{32.4} & \textbf{35.8} & \textbf{39.4} & \textbf{44.0} & \textbf{38.9} & \textbf{59.0} & 21.6 \\
\midrule
Qwen2.5-14B-Instruct & 35.3 & 34.9 & 35.5 & 43.3 & 32.6 & 27.1 & 33.8 & 33.3 & 32.0 & 38.3 & 35.9 & 33.6 \\
R1-Distill-Qwen-14B  & 36.2 & 40.2 & 33.8 & 44.1 & 31.4 & 32.6 & 36.8 & 36.4 & 28.4 & 38.4 & 44.1 & 33.4 \\
R1-Distill-Qwen-32B & 38.6 & 40.1 & 37.6 & 48.9 & 33.5 & 31.5 & 29.6 & 39.4 & 38.0 & 39.4 & 51.3 & 39.2 \\
QwenLong-L1-32B & 40.8 & \bf 46.4 & 37.4 &\bf 52.4 & 35.7 & 31.5 & 37.0 & 32.7 & \bf 43.2 & 40.0 & 55.4 & \bf 41.0 \\
\textbf{LoongRL-14B} &\bf 42.3 &\bf 46.4 &\bf  39.9 & 44.4 & \bf 43.3 & \bf 37.0 &\bf  39.5 & \underline{\textbf{45.5}} & 38.0 & \textbf{44.0} & \textbf{59.0} & 37.6 \\
\bottomrule
\end{tabular}
}
\label{tab:lb2}
\end{table*}

We evaluate the LoongRL series and baseline models on the LongBench-v2 benchmark. 
For models without long chain-of-thought (Long CoT) reasoning capability, we follow the original LongBench-v2 CoT setting, using a temperature of $0.1$, sampling five responses per query, and reporting the average score across them (\textit{Avg@5}). 
For models with Long CoT reasoning ability, we instead adopt a temperature of $0.6$, again sampling five responses and reporting \textit{Avg@5}. 
In addition, for the Qwen family and our LoongRL models, we apply the YaRN method to extend the context length to $128$k tokens.
The overall comparison results are summarized in Table~\ref{tab:lb2}.

\subsection{RULER}
We evaluated the retrieval capabilities of models on long-text tasks using the RULER benchmark.
For models without long-context reasoning abilities, we followed the original RULER setting by appending a prompt suffix designed to guide the model to produce completion-style answers, e.g. ``What is the special magic number for wandering-age mentioned in the provided text? The special magic number for wandering-age mentioned in the provided text is''.
In contrast, for models capable of long-context reasoning, we removed this completion-style suffix, as preliminary experiments indicated that these models tend not to provide direct completions but instead perform explicit reasoning before answering. 
After removing the suffix, we allowed the reasoning-capable models to generate up to 8192 tokens and subsequently extracted the model’s answer from the text following the ``\texttt{</think>}'' token in its output.

\begin{table}[t!]
\small
\centering
\caption{RULER benchmark results across different context lengths. For QwQ, QwenLong, Qwen2.5 model series, we report their YaRN variants for 64k and 128k.}
\label{tab:ruler_results}
\setlength{\tabcolsep}{5pt}
\begin{tabular}{lccccccg}
\toprule
\textbf{Model} & \textbf{4k} & \textbf{8k} & \textbf{16k} & \textbf{32k} & \textbf{64k} & \textbf{128k} & \textbf{Avg.} \\
\midrule
o3-mini & 96.58 & 96.85 & 94.69 & 90.85 & 74.81 & 65.40 & 86.53 \\
DeepSeek-R1 & 98.46 & 97.98 & 97.18 & 96.06 & 94.92 & 85.10 & 94.95 \\
GPT-4o & 97.69 & 96.73 & 96.73 & 96.02 & 94.46 & 89.10 & 95.12 \\
QwQ-32B (YaRN@64k/128k) & 89.10 & 86.46 & 83.84 & 78.42 & 64.72 & 59.68 & 77.37 \\
R1-Distill-LLaMa-70B & 94.89 & 95.60 & 93.75 & 89.60 & 79.65 & 0.00 & 75.58 \\
Llama3.1-70B-Instruct & 96.78 & 96.64 & 95.82 & 94.87 & 89.21 & 64.53 & 89.64 \\
\midrule
R1-Distill-LLaMa-8B & 83.89 & 79.80 & 73.77 & 64.46 & 51.06 & 1.28 & 59.04 \\
Llama3.1-8B-Instruct & 96.10 & 93.81 & 90.91 & 86.73 & 84.77 & 74.15 & 87.75 \\
Qwen2.5-7B-Instruct (YaRN@64k/128k) &\bf 95.16 & 93.73 & 92.31 & 89.46 & 81.79 & 69.41 & 86.31 \\
R1-Distill-Qwen-7B & 65.70 & 48.29 & 18.86 & 4.38 & 1.41 & 0.88 & 23.25 \\
LoongRL-7B (YaRN@64k/128k) & 95.06 &\bf 94.34 & \bf93.37 & \bf91.36 & \bf86.18 & \bf76.84 & \bf89.53 \\
\midrule
Qwen2.5-14B-Instruct (YaRN@64k/128k) & 96.27 & 95.11 & 93.38 & 92.53 & 82.33 & 73.57 & 88.86 \\
R1-Distill-Qwen-14B & 91.44 & 86.29 & 85.73 & 82.00 & 60.24 & 28.23 & 72.32 \\
R1-Distill-Qwen-32B & 93.61 & 91.64 & 90.27 & 88.90 & 71.51 & 40.88 & 79.47 \\
QwenLong-L1-32B (YaRN@64k/128k) & 91.71 & 88.51 & 87.55 & 86.81 & 80.64 & 70.19 & 84.24 \\
LoongRL-14B (YaRN@64k/128k) & \bf97.56 & \bf96.14 & \bf95.36 & \bf95.11 & \bf87.14 & \bf79.92 &\bf 91.87 \\
\bottomrule
\end{tabular}
\end{table}

\subsection{Needle-in-a-Haystack}
\begin{figure}[ht]
\centering
\includegraphics[width=0.6\linewidth]{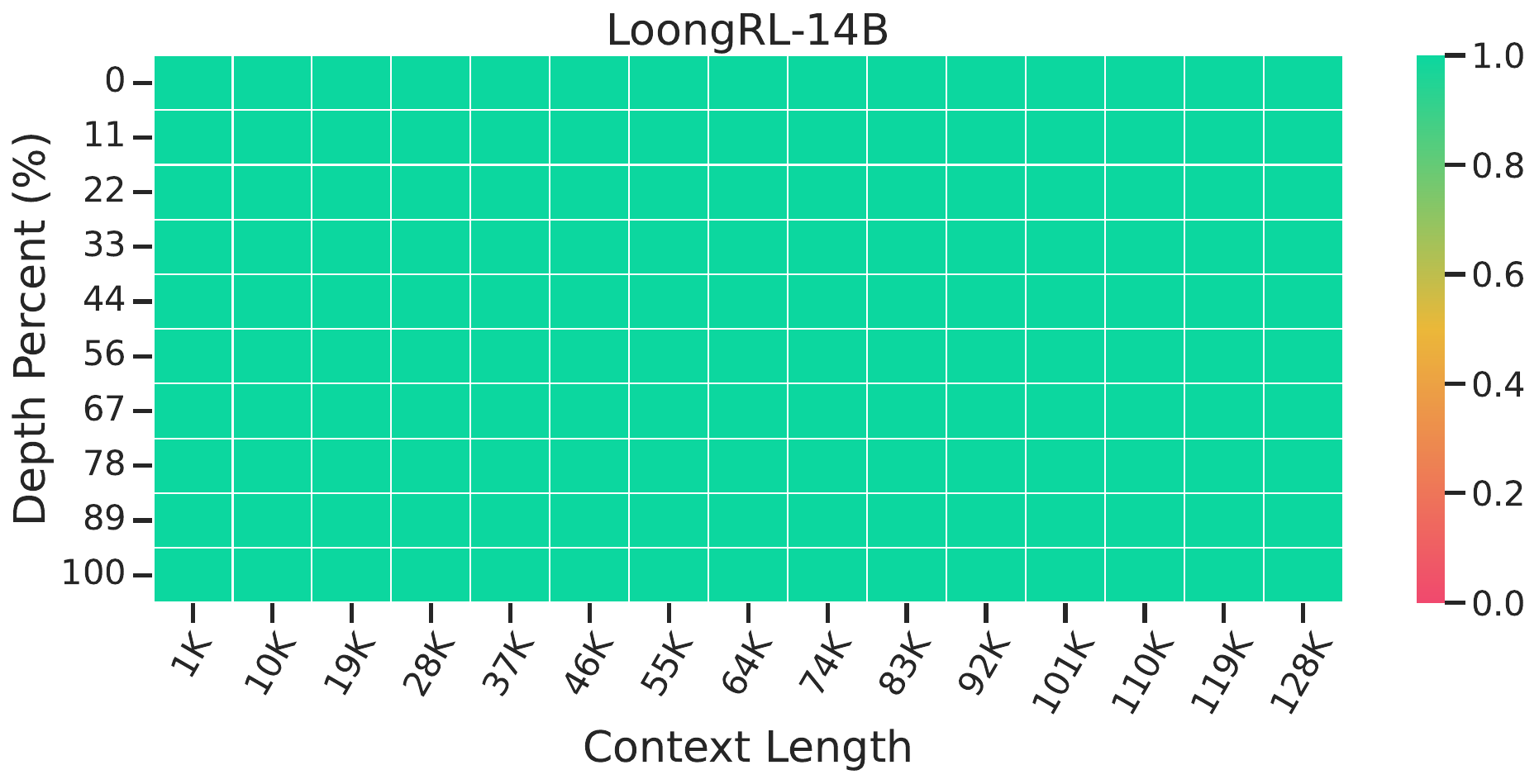}
\caption{Needle-in-a-Haystack performance of LoongRL-14B.}
\label{fig:niah_14b}
\end{figure}

We further evaluated the \textit{needle-in-a-haystack} (NIAH) task, 
which specifically measures the retrieval ability of models in extremely long-text settings. 
Figure~\ref{fig:niah_14b} reports the performance of our LoongRL-14B model. 
Results demonstrate that LoongRL-14B maintains strong retrieval accuracy across extended context lengths, 
showcasing its robustness for long-context information retrieval.

\end{document}